\documentclass[journal,12pt,onecolumn,draftclsnofoot,]{IEEEtran}
\usepackage{graphicx}
\ifCLASSINFOpdf
\else
\fi
\usepackage{amsmath}
\usepackage{algorithmic}
\usepackage{array}
\usepackage{tikz}
\usetikzlibrary{shapes,arrows}
\usepackage{schemabloc}

\usetikzlibrary{circuits}
\usepackage{tikz}
\usetikzlibrary{shapes,arrows,chains}
\usetikzlibrary{shapes.geometric, arrows}
\usetikzlibrary{shapes,snakes}
\usepackage[latin1]{inputenc}
\usetikzlibrary{calc,patterns}
\usetikzlibrary{quotes,angles}
\usetikzlibrary{shapes,arrows}
\usetikzlibrary{automata, positioning}
\usetikzlibrary{arrows.meta}
\tikzset{%
  >={Latex[width=2mm,length=2mm]},
            base/.style = {rectangle, rounded corners, draw=black,
                           minimum width=4cm, minimum height=1cm,
                           text centered, font=\sffamily},
  activityStarts/.style = {base, fill=blue!30},
       startstop/.style = {base, fill=red!30},
    activityRuns/.style = {base, fill=green!30},
         process/.style = {base, minimum width=2.5cm, fill=orange!15},
}

\tikzset{
block/.style={
  draw, 
  fill=blue!20, 
  rectangle, 
  minimum height=3em, 
  minimum width=3em
  },
  block1/.style={
  draw, 
  fill=blue!20, 
  rectangle, 
  minimum height=3em, 
  minimum width=2em
  },
sum/.style={
  draw, 
  fill=blue!20, 
  circle, 
  },
input/.style={coordinate},
output/.style={coordinate},
pinstyle/.style={
  pin edge={to-,thin,black}
  }
} 

\begin{document}
%\onecolumn
\title{Integrated guidance and control framework for the waypoint navigation of a  miniature aircraft with highly coupled longitudinal and lateral dynamics}

\author{K Harikumar, Jinraj V. Pushpangathan,
        Sidhant Dhall,
        and M. Seetharama Bhat 
\thanks{K. Harikumar is a Research Fellow @  Singapore University of Technology and Design, Singapore,
 e-mail: harikumar100@gmail.com, }% <-this % stops a space
 \thanks{Jinraj V. Pushpangathan is a Research Fellow @ Department of Aerospace Engineering, Indian Institute of Science, Bangalore, India,  e-mail: jinrajaero@gmail.com }
\thanks{Sidhant Dhall is a Research Assistant @ Department of Aerospace Engineering, Indian Institute of Science, Bangalore, India, email: sidhant.dhall@gmail.com }% <-this % stops a space
\thanks{M. Seetharama Bhat is a Professor @ Department of Aerospace Engineering, Indian Institute of Science, Bangalore, India,  e-mail: msbdcl@aero.iisc.ernet.in}}


\markboth{IEEE}%
{Shell \MakeLowercase{\textit{et al.}}: Bare Demo of IEEEtran.cls for IEEE Journals}



\maketitle

\begin{abstract}
\noindent A solution to the waypoint navigation problem for fixed wing micro air vehicles (MAV) is addressed in this paper, in the framework of integrated guidance and control (IGC). IGC yields a single step solution to the waypoint navigation problem, unlike conventional multiple loop design. The  pure proportional navigation (PPN) guidance law is integrated with the  MAV dynamics. A multivariable static output feedback (SOF) controller is designed for the linear state space model formulated in IGC framework. The waypoint navigation algorithm handles the minimum turn radius constraint of the MAV. The algorithm also evaluates the feasibility of reaching a waypoint. Extensive non-linear simulations  are performed on high fidelity 150 mm wingspan MAV model to demonstrate the potential advantages of the proposed waypoint navigation algorithm.
\end{abstract}


% \fi
\IEEEpeerreviewmaketitle

\section{Introduction}
\noindent  Fixed wing micro air vehicles (MAV) are extremely agile, lightweight aircraft with a maximum wingspan of 150 mm \cite{mav1}. In any of the MAV applications, it has to reach a set of waypoints, specified by the end user. In most of the waypoint navigation problems, guidance and control are treated as two separate problems. Control system is usually designed to give higher bandwidth to track the commands generated by the guidance law. But when the guidance law is integrated with the inner loop control system, the combined system often fails to meet the required performance and stability specifications. The major reasons are due to the  assumption of lower order dynamics for the inner loop control system, neglecting actuator bandwidth and assuming decoupled longitudinal and lateral dynamics, especially for the case of MAVs. Many of the guidance law in the literature assumes lower order autopilot for inner loop control system \cite{Mclain}. The assumption of lower order inner loop  autopilot provides fictitious high amplitude gain margin for guidance law to operate.  When the combined system is not stable, the guidance law parameters are retuned to ensure stability. The re-tuning of guidance law parameters often lead to inadequate performance. The actuator bandwidth is of major concern for MAVs. The short period mode and Dutch roll mode natural frequencies of MAVs are comparable to actuator bandwidth \cite{hari1}. The relatively lower actuator bandwidth in MAVs leads to poor wind disturbance rejection. The wind disturbances change angle of attack $(\alpha )$ and side slip angle $(\beta)$ considerably from the trim operating conditions. The additional forces and moments generated by wind disturbances are often neglected while designing the guidance law in the presence of wind \cite{shika}.  The coupling between longitudinal and lateral dynamics is severe for MAVs and cannot be neglected as in the case of bigger UAVs \cite{hari1},\cite{hari2}. The guidance law design needs to consider the combined longitudinal and lateral dynamics, failure of which leads to instability during abrupt maneuvers. \\
\noindent The topic of integrated guidance and control (IGC) has gained considerable interest in the area of missile guidance and control \cite{Hyan}-\cite{ast1} due to the following advantages. The synergy between the guidance and control subsystems can be exploited to make the combined system more stable and effective than the individual subsystems \cite{Hyan}. Despite the fact that separate guidance and control has been applied successfully,  still there remains scope for further optimizing the performance and enhancing the stability \cite{Mxin}. The prime focus on the missile IGC design is to achieve improved terminal miss distances. When the missile approaches the target, the time scale separation between the control loop and guidance loop is no longer valid \cite{Yike}.  The conventional approach of designing separate guidance and control leads to a conservative design with the requirement of actuators with higher bandwidth than necessary \cite{menon}.  \\
\noindent  The idea of IGC is also applied to fixed wing UAVs \cite{kaminer}-\cite{babar}. A linear time invariant model in the framework of integrated guidance and control is developed for a fixed wing UAV of wingspan 3.8 $m$ to track helical trim trajectories in \cite{kaminer}. For a UAV flying along the trim trajectory, the  net forces and moments acting on the UAV are zero. The method developed in \cite{kaminer} explains how to follow a given helical trajectory in which UAV remains in a trimmed state. But finding such a trimmed trajectory connecting two waypoints is very computationally intensive. Moreover, the method cannot be applied for in-flight waypoint changes since it is difficult to compute the trim trajectories online with the limited onboard computational power available with MAV. The IGC design for a fixed wing UAV of 8.8 $m$ wingspan for following a predefined path using sliding mode control is explained in \cite{yamasaki}. The design assumes that the roll, pitch and yaw dynamics are decoupled. But for  MAVs, the roll, pitch and yaw dynamics are tightly coupled \cite{hari2}. Nonlinear dynamic inversion (NDI) based IGC framework is developed for UAVs in \cite{padhi} and \cite{xliu}.  NDI utilizes full state feedback structure. In the case of MAVs, measurement of $\alpha$ and $\beta$  is tedious due to the lack of availability of lightweight sensors. Moreover, the perturbations in $\alpha$ and $\beta$ are very high for the MAVs when compared to bigger UAVs under nominal wind conditions.  IGC framework developed for lateral dynamics of a fixed wing UAV  assumes decoupled longitudinal dynamics in \cite{babar}. But the assumptions of decoupled longitudinal and lateral dynamics is not valid for 150 mm class of MAVs \cite{hari1},\cite{hari2}. \\
To overcome the above-mentioned limitations of existing IGC methodology, a novel IGC framework is derived in this paper, combining the idea of pure proportional navigation (PPN) with MAV dynamics. The proposed waypoint navigation algorithm  based on IGC framework  is non-iterative and hence computationally inexpensive. The  issue of coupling between longitudinal and lateral dynamics  of the MAV is handled by the proposed IGC framework. Unlike the existing algorithms based on IGC framework, the proposed waypoint navigation algorithm  evaluates the feasibility of reaching a waypoint based on the acceleration constraints of the MAV. A $H_{\infty}$  SOF controller is designed for the combined guidance and control loop using Linear Matrix Inequality (LMI) techniques. Numerical simulation results are presented for the straight line following and rectangular path following using high fidelity nonlinear model of a MAV obtained from wind tunnel data.\\
The paper is organized as follows. The coupled linear state space model of the MAV is presented in section \ref{coupling}. The guidance law used in IGC framework is given in section \ref{guidance}. The integration of guidance law to the coupled dynamics of MAV and the waypoint navigation algorithm is presented in section \ref{igc}. The non-linear six degrees of freedom  simulations for straight line following and rectangular path following is given in section \ref{numsim}, followed by conclusions.

% The main differences between separate guidance and control and integrated guidance and control is summarized in Table \ref{compigc}.
%\begin{table}[!ht]
% \begin{center}
% \caption{Differences between separate guidance and control and IGC}
%\begin{tabular}{|p{7cm}|p{7cm}|} % 11 cols (lef, right); vert. lines
%\hline % draw horizontal line
%\textbf{Separate guidance and control}  &  \textbf {Integrated guidance and control}\\
%\hline
%Assumes decoupled lateral and longitudinal
%dynamics. This may lead to poor performance and instability during abrupt maneuvers  &  Both longitudinal and lateral dynamics are integrated in the design  \\
%\hline
%  Intensive tuning of guidance parameters is required for achieving a good practical solution  & Tuning is not required as the full dynamics is incorporated   \\
%\hline
% Higher control system bandwidth  is required for good performance  & No separate restrictions on  control system bandwidth   \\
%\hline
% Assumes lower order autopilot for the commanded variables. This assumption may generate unrealizable guidance commands  & Full order of the autopilot is taken into account \\
%\hline
%\end{tabular}
%\label{compigc}
%\end{center}
%%\end{table}
%\IEEEPARstart{T}{his} demo file is intended to serve as a ``starter file''
%for IEEE journal papers produced under \LaTeX\ using
%IEEEtran.cls version 1.8b and later.
%% You must have at least 2 lines in the paragraph with the drop letter
%% (should never be an issue)
%I wish you the best of success.
\section{Preliminaries} {\label{coupling}}
\subsection{Coupled linear state space model of the MAV} 
\noindent  The specifications of the MAV considered in this paper is given in Table \ref{mavspec}, \cite{mav1}, \cite{hari2}.
\begin{table}[!ht]
\caption{Specifications of the MAV}
\centering
\begin{tabular}{|c|c|}
\hline
Planform & Rectangular \\
\hline
Take-off weight & 53 grams \\
\hline
Cruise airspeed & 8 $m/s$\\
\hline
Stall airspeed & 6 $m/s$\\
\hline
Wing - chord length  & 0.11 $m$\\
\hline
Wing - span length  & 0.15 $m$ \\
\hline
Airfoil & E387 \\
\hline
Control surfaces & Elevator and rudder \\
\hline
\end{tabular} \\
\label{mavspec}
\end{table}

\noindent The coupling between longitudinal and lateral dynamics is severe for the case of MAVs. The coupled linear state space shows an unstable spiral mode, whereas the decoupled model fails to capture the unstable spiral mode \cite{hari2}. The effects of asymmetric propeller wake, motor-propeller counter torque and gyroscopic effects lead to unstable spiral mode dynamics \cite{hari2}, \cite{jj}. The coupled linear state space model of the MAV is given in (\ref{Lintable}). 
\begin{equation}
\dot{X}=A_cX+B_cU
\label{Lintable}    
\end{equation}
where 
\begin{eqnarray}
X=[\widetilde{ u }, \widetilde{ w }, \widetilde{ q }, \widetilde{ \theta }, \widetilde{ h }, \widetilde{ v }, \widetilde{ p } ,\widetilde{ r },\widetilde{ \phi }]^T
 \label{Lintable1}
\end{eqnarray}
 and 
 \begin{equation}
  U=[\widetilde{\delta}_{es},\widetilde{\delta}_{rs},\widetilde{\delta_{t}}]^T 
  \label{Lintable2}
 \end{equation}
 
 \noindent The set ($\widetilde{ u },\widetilde{ v },\widetilde{ w }$) represents the linearized velocity along body axis, ($\widetilde{ p },\widetilde{ q },\widetilde{ r}$) are the linearized angular rates, $\widetilde{ \phi }$ is the linearized roll angle, $\widetilde{ \theta }$ is the linearized pitch angle  and  $\widetilde{ h }$ is the linearized altitude. The linearized elevator servo output $\widetilde{\delta}_{es}$, linearized rudder servo output $\widetilde{\delta}_{rs}$ and the linearized thrust generated by motor-propeller denoted by $\widetilde{\delta_{t}}$ are the control inputs. $A_{c}$ is the system matrix for coupled state space model and $B_{c}$ is the input matrix. The  matrices $A_c$ and $B_c$ are given in \cite{hari2} for the case of 150 mm wingspan fixed wing MAV. 

%             \begin{equation}
% % \begin{split}
% \left(
% \begin{array}{c}
% \dot {\widetilde{ u }} \\
% \dot{\widetilde{ w }}\\
% \dot {\widetilde{ q}}\\
% \dot {\widetilde{ \theta }}\\
% \dot {\widetilde{ h }}\\
%  \dot {\widetilde{v }} \\
% \dot {\widetilde{ p }}\\
% \dot {\widetilde{ r}}\\
% \dot {\widetilde{ \phi }}
% \end{array}
% \right)=  A_{c}    \left(
% \begin{array}{c}
%  {\widetilde{ u }}\\
%  {\widetilde{ w }}\\
%  {\widetilde{ q}}\\
%  {\widetilde{ \theta }}\\
%  {\widetilde{ h}}\\
%  {\widetilde{v }}\\
% {\widetilde{ p }}\\
%   {\widetilde{ r}}\\
% {\widetilde{ \phi }}
% \end{array}
% \right)
% \\+
% B_{c} 
% \left(
% \begin{array}{c}
% \widetilde{ \delta_{es}}\\
%  \widetilde{ \delta_{rs}}\\
% \widetilde{ \delta_{t}}
% \end{array}
% \right)
% % \end{split}
% \label{Lintable}
% \end{equation}

\noindent The states of longitudinal dynamics is given by
\begin{equation}
X_1=[\widetilde{ u }, \,\, \widetilde{ w },\,\, \widetilde{ q},\,\, \widetilde{ \theta },\,\, \widetilde{ h }]^T 
\label{state1a}
 \end{equation}
 Similarly, the states of lateral dynamics is given by
\begin{equation}
X_2=[\widetilde{v },\,\widetilde{ p },\,\, \widetilde{ r},\,\, \widetilde{\phi}]^T 
\label{state1b}
 \end{equation}
 The  matrix  $A_c$ in (\ref{Lintable}) can be written as
\begin{equation}
{ A_c}
=\left(\begin {array}{cc} A_{11} & A_{12}\\    A_{21} & A_{22}\end {array} \right)
\label{acoup1}
\end{equation} 
where $A_{11}$ denotes the state matrix for decoupled longitudinal dynamics and $A_{22}$ the state matrix for decoupled lateral dynamics. The coupling between longitudinal and lateral dynamics is introduced by the matrices $A_{12}$ and $A_{21}$.
For bigger UAVs, the eigenvalues of $A_c$  denoted by $\bar{\lambda}(A_c)$ = $\bar{\lambda}(A_{11})$ $\cup$  $\bar{\lambda}(A_{22})$. But for the case of 150 mm wingspan MAVs,
$\bar{\lambda}(A_c)$ $\neq$ $\bar{\lambda}(A_{11})$ $\cup$  $\bar{\lambda}(A_{22})$ \cite{hari2}.\\
\textit{Remark}: In the rest of this paper, a variable with $\tilde{.}$ symbol denotes the linearized variable. For example, the variables ($u,\,,v,\,w$) denotes the velocity components along the body axis and ($\widetilde{ u },\,\widetilde{ v },\,\widetilde{ w }$) represents their linearized version.

\section{Guidance logic  for  waypoint navigation in two dimensional space} 
\begin{figure}[!h] 
                 \begin{center}
                  \includegraphics[trim = 0cm 0cm 0cm 0cm, clip=true, scale=0.60]{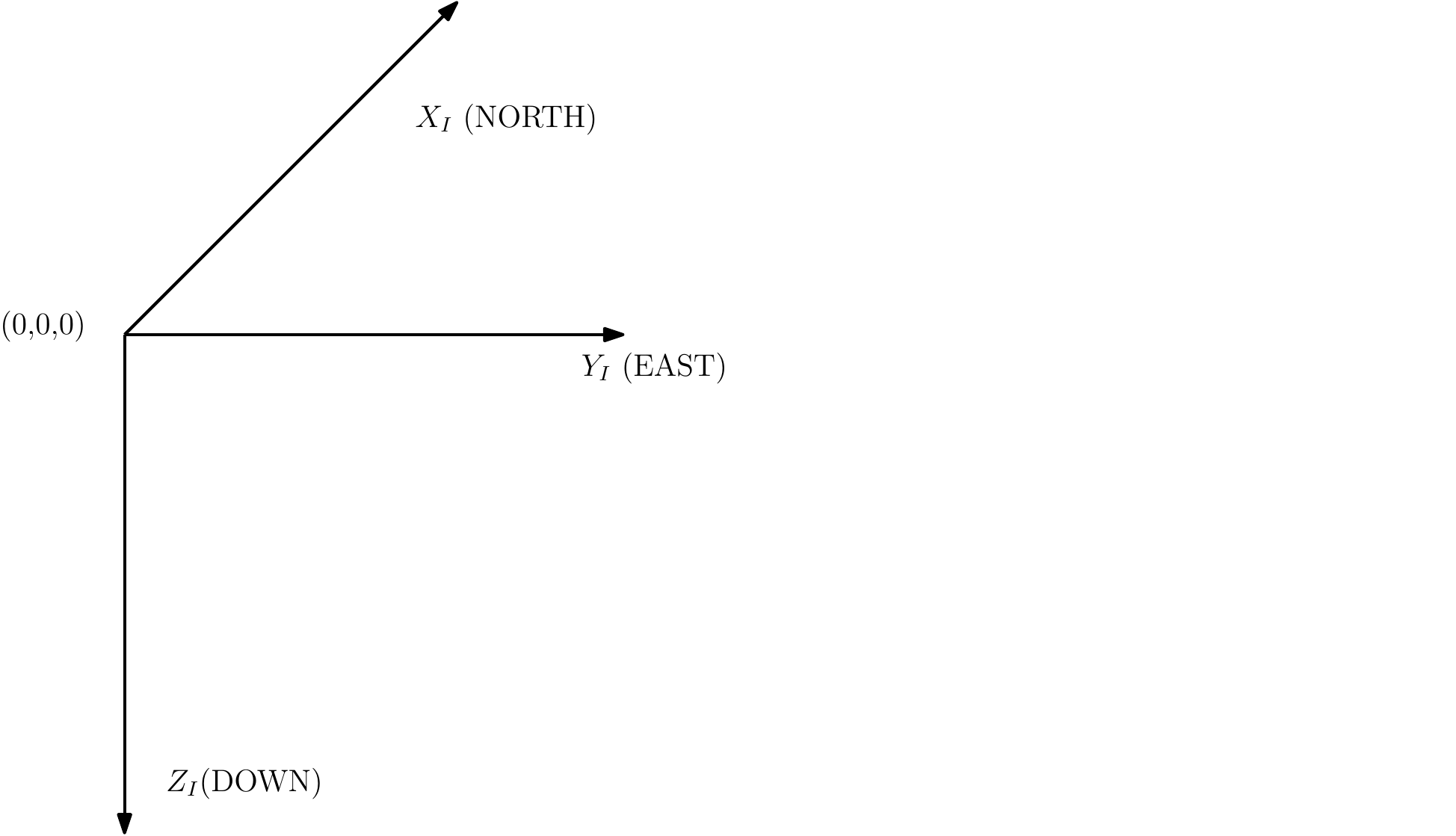}
                 \caption{Diagram showing inertial $X_{I}Y_{I}Z_{I}$ co-ordinate sytem}
                 \label{fig1}
                 \end{center}
                 \end{figure}

\begin{figure}[!h] 
                 \begin{center}
                  \includegraphics[trim = 0cm 0cm 0cm 0cm, clip=true, scale=0.60]{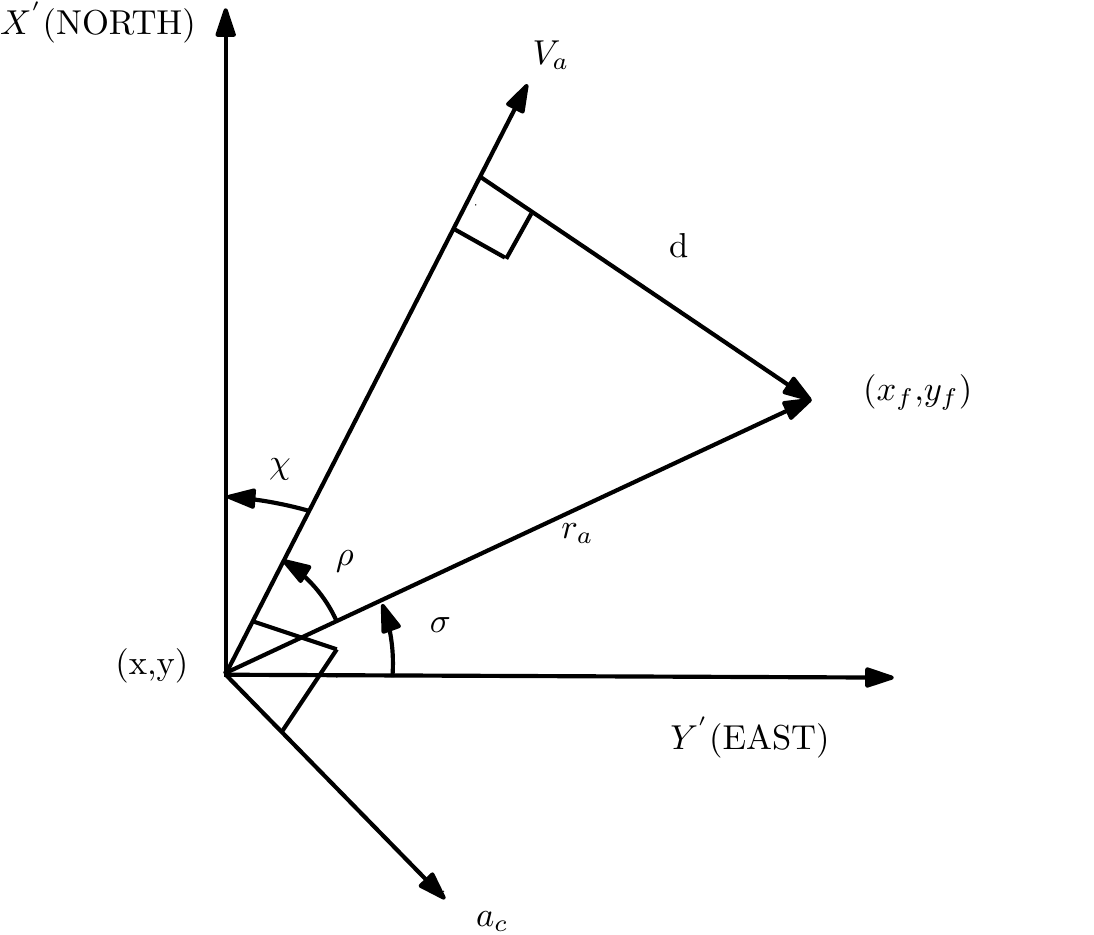}
                 \caption{Diagram explaining the guidance law}
                 \label{3D3}
                 \end{center}
                 \end{figure}

{\label{guidance}}
\noindent The origin of the local inertial coordinate system ($X_{I}Y_{I}Z_{I}$) is  the home location of the MAV as shown in Fig.  \ref{fig1} with positive $X_{I}$ axis  towards the  geographic north pole, positive $Y_{I}$ axis in the direction of  $90^o$ clockwise  rotation from positive $X_{I}$ axis and  positive $Z_{I}$ axis points towards the down direction. For the waypoint following in 2D space, altitude hold is employed using altimeter feedback, so motion in the local inertial $X_{I}Y_{I}$ plane is only considered.  The guidance logic  is explained first in this section and the integration of the guidance logic with the control system is explained later in section IV.\\
\noindent In Fig. \ref{3D3},  axes $X'Y'$ are parallel to $X_{I}Y_{I}$, $(x,y)$ is the  current MAV position in the $X_{I}Y_{I}$ plane, $(x_{f},y_{f})$ is the next waypoint to be followed, $r_{a}$ is the range, $\sigma$ is the angle between the line joining  $(x,y)$ ,  $(x_{f},y_{f})$ and the  $Y'$ axis, $\chi$ is the angle made by the velocity vector with the $X'$ axis. The miss distance  $d$ is the perpendicular distance between $(x_{f},y_{f})$ and the velocity vector.  The  applied acceleration  is  denoted by $a_{c}$ and is perpendicular to the velocity vector and lies in the $X_{I}Y_{I}$ plane. The units of $(x,y)$, $(x_{f},y_{f})$, $d$, $r_{a}$ are in meters (m), $V_{a}$ is in meters/second ($m/s$), $a_{c}$ is in meters/second-square($m/s^2$) and $\sigma$,  $\chi$ are in radians(rad). From Fig. \ref{3D3}, we can write the following relations given in (\ref{sigma1}) to (\ref{gdlaw}).

  \begin{eqnarray}
 \sigma=tan^{-1} (\frac{x_{f}-x}{y_{f}-y})
\label{sigma1}
\end{eqnarray}
  \begin{eqnarray}
 \rho=\frac{\pi}{2}-(\chi+\sigma)
\label{rho1}
\end{eqnarray}
  \begin{eqnarray}
r_{a}=\sqrt{(x_{f}-x)^{2}+(y_{f}-y)^{2}}
\label{ra1}
 \end{eqnarray}
  \begin{eqnarray}
  d=r_{a}sin\rho
 \label{Glawd}
 \end{eqnarray}
  \begin{eqnarray}
 a_{c}=k_{1}d
\label{gdlaw}
 \end{eqnarray}

\noindent In (\ref{gdlaw}), $k_1$ is the proportional gain. The magnitude of applied acceleration  depends upon $k_1$. An expression for the gain $k_{1}$ is given in the subsection \ref{ppnguidelaw}. The  acceleration is applied perpendicular to the velocity vector in such a way that it does not change the magnitude of velocity but only the direction. This is analogues to  Pure Proportional Navigation (PPN) where the acceleration applied is perpendicular to the  velocity vector \cite{shneydor}.  Here it is assumed that altitude and $V_{a}$ is constant. When the miss distance is zero, from (\ref{Glawd}), we  obtain either  $r_{a}=0$ or $\rho=0$. The condition  $r_{a}=0$  implies that $x=x_{f}$ and   $y=y_{f}$. From  Fig. \ref{3D3} , $\rho=0$ implies that the velocity vector is pointing towards the goal point. So when the miss distance is zero, the MAV has either reached the desired waypoint or the velocity vector of the MAV is along the line connecting the waypoint  and current MAV position such that no further acceleration has to be applied to change the direction of the MAV. 

\subsection{Relation to PPN Guidance Law and choice of gain $k_1$} \label{ppnguidelaw}
\begin{figure}[!h] 
                 \begin{center}
                 \includegraphics[trim = 0cm 0cm 0cm 0cm, clip=true, scale=0.60]{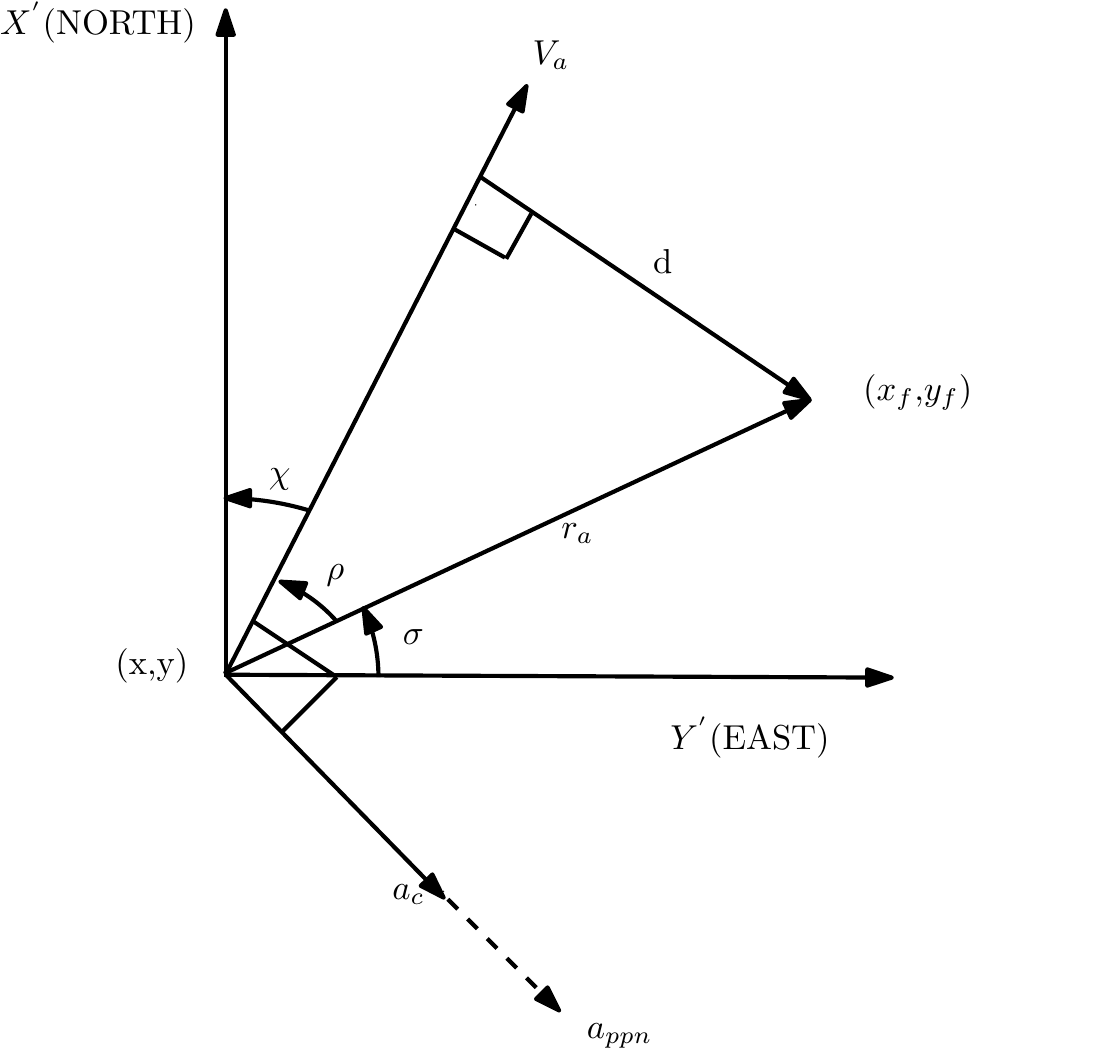}
 \caption{Diagram used for explaining the relation between the proposed guidance law and PPN guidance law}
                 \label{3D23}
                 \end{center}
                 \end{figure}

\noindent This subsection gives the relation between the proposed guidance law given in    (\ref{gdlaw}) and the  Pure Proportional Navigation (PPN) guidance law. Consider Fig. \ref{3D23}, in which  $a_{ppn}$  denotes the acceleration applied in PPN guidance law and rest all variables are same as in Fig. \ref{3D3}. 
The expression for  PPN guidance law \cite{shneydor} is given by
\begin{eqnarray}
a_{ppn}=-NV_{a}\dot{\sigma}
\label{PN}
\end{eqnarray}
where $\dot{\sigma}$ is the line of sight rate and $N$ is the navigation constant.
%where $a_{ppn}$ is the acceleration perpendicular to the velocity vector, $V_{c}$ is the closing velocity and is given by equation \ref{closevel}, $\dot{\sigma}$ is the line of sight rate and $N$ is the navigation constant.
%\begin{eqnarray}
%V_{c}=-V_{a}\,cos{\rho}
%\label{closevel}
%\end{eqnarray}
\begin{eqnarray}
\dot{\sigma}=-\frac {V_{a}\,sin{\rho }}{r_a}
\label{losr}
\end{eqnarray}
Using (\ref{PN}) to (\ref{losr}),
\begin{eqnarray}
a_{ppn}=NV_{a}( \frac{V_{a}\,sin\rho}{r_a})
\label{alos1}
\end{eqnarray}
%Since $V_{a}$ is constant,
%\begin{eqnarray}
%a_{ppn}=N_{1} \frac{{{cos(\rho)}^2}\,sin(\rho)}{r_a}
%\label{alospn}
%\end{eqnarray}
%where $N_{1}=N {V_{a}}^2$. 
%For the guidance law given in equation \ref{gdlaw}, let the acceleration component along the direction of  $a_{Lospn}$ be denoted by $a_{Los}$. Then 
%\begin{eqnarray}
%a_{Los}=a_{c}sin(\chi+\sigma)
%\label{alos1}
%\end{eqnarray}
From (\ref{Glawd}) and (\ref{gdlaw}),
\begin{eqnarray}
a_{c}=k_{1}r_{a}sin\rho
%a_{c}=k_{1}r_{a}sin(\frac{\pi}{2}-(\chi+\sigma))\\
%a_{c}=k_{1}r_{a}cos(\chi+\sigma)
\label{alos2}
\end{eqnarray}
Substituting (\ref{alos2}) into (\ref{alos1}) gives,
\begin{eqnarray}
a_{ppn}=(\frac{N}{k_1}){(\frac{V_a }{r_a})^2}a_c
\label{alos3}
\end{eqnarray}
%From equations \ref{alospn} and  \ref{alos3},
%\begin{eqnarray}
%a_{Los}=k_{1}r_{a}(\frac{a_{Lospn}r_{a}}{N_{1}})\\
%a_{Los}=(\frac{k_{1}{{r_{a}}^2}}{N_{1}})a_{Lospn}
%\label{alos4}
%\end{eqnarray}
%Thus the proposed guidance law is a scaled version of the PN guidance law with a scaling factor of $\frac{k_{1}{{r_{a}}^2}}{N_{1}}$.  Substituting $N_{1}=N {V_{a}}^2$, equation \ref{alos4} becomes,
%\begin{eqnarray}
%a_{Los}=(\frac{k_{1}{{r_{a}}^2}}{N{V_{a}}^2})a_{Lospn}\\
%\label{alos5}
%\end{eqnarray}
If we choose  $k_{1}$ as given in   (\ref{alos6}), the proposed guidance law is same as the PPN guidance law and is optimal for N=3 as shown in \cite{yansh}.
\begin{eqnarray}
k_{1}=\frac{N(V_{a})^2}{{r_{a}}^2}
\label{alos6}
\end{eqnarray}
Thus the guidance law given in (\ref{gdlaw}) takes the form 
\begin{eqnarray}
a_{c}=(\frac{N({V_{a}})^2}{{r_{a}}^2})d
\label{gdlawf}
\end{eqnarray}
and is same as the PPN guidance law.
The guidance law given in (\ref{gdlawf}) will ensure that the MAV will reach the waypoint for $N>=2$, \cite{shneydor}. 
\subsection{Constraint on minimum turn radius }
\noindent Micro air vehicles have a constraint on the minimum allowable turn radius. The minimum value of the turn radius denoted by $R_{min}$ depends upon the airspeed, stall angle of attack and the maximum thrust available.  Since the applied acceleration is perpendicular to the velocity vector, the MAV follows a circular path with an instantaneous radius of curvature inversely proportional to the magnitude of the applied acceleration. The constraint on the applied acceleration $a_c$ can be given as \cite{meriam}
\begin{eqnarray}
a_c < \frac{{V_a}^2}{R_{min}}
\label{rcons1}
\end{eqnarray}
Using (\ref{gdlawf}) and (\ref{Glawd}),   (\ref{rcons1}) can be written as
\begin{eqnarray}
(\frac{N(V_{a})^2}{{r_{a}}^2})(r_a sin\rho)< \frac{{V_a}^2}{R_{min}}
\label{rcons2}
\end{eqnarray}
After simplifying, (\ref{rcons2}) becomes
\begin{eqnarray}
\frac{Nsin\rho}{r_{a}}< \frac{1}{R_{min}}
\label{rcons3}
\end{eqnarray}
The minimum value for $N=2$. Hence   (\ref{rcons3}) becomes,
% \begin{eqnarray}
% N< \frac{r_{a}}{R_{min}}
% \label{rcons4}
% \end{eqnarray}
\begin{eqnarray}
r_a>2 R_{min} sin\rho
\label{rcons4}
\end{eqnarray}
At any instant, the relation given by  (\ref{rcons4}) ensures that the minimum turn radius constraint is not violated.  A waypoint is considered to be not feasible if the condition given in  (\ref{rcons4}) is violated.\\
\section{Integrated guidance and control  for waypoint navigation in two dimensional space} 
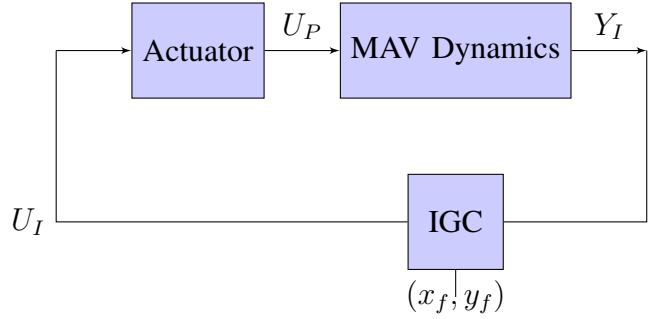
\begin{figure}[!h]
    \centering
  \begin{tikzpicture}[auto,>=latex']
    % We start by placing the blocks
    \node [input, name=input] {};
   % \node [sum, right = of input] (sum1) {};
    \node [block, right = of input] (pac) {Actuator};
    %\node [sum, right = of pac] (sum2) {};
    \node [block, right = of pac] (controller) {MAV Dynamics};
   % \node [block, right = of controller
     %       ] (system) {Sensor $\&$ Navigation};
    % We draw an edge between the controller and system block to 
    % calculate the coordinate u. We need it to place the measurement block. 
    %\draw [->] (controller) -- node[name=u] {} (system);
    \node [output, right =of controller] (output) {$Y_I$};
     \node [block, below = of controller
            ] (system1) {IGC};
    \node [output, right =of system1] (output1) {};
    % \node [, below =of controller] (a1) {$y_1(k)$};
\node [output, right = of system1](way) {};
\draw [draw,-] (system1) -- ++(0.0 cm,-1.0cm) node {$(x_f,y_f)$} (way);
    % Once the nodes are placed, connecting them is easy. 
    \draw [draw,->] (input) -- node {} (pac);
    %\draw [draw,->] (sum1) -- node {$e_{\psi}$} (pac);
   %\draw [draw,->] (pac) -- node {$r_d$} (sum2);
    \draw [->] (pac) -- node {$U_P$} (controller);
    \draw [->] (controller) -- node [name=y] {$Y_I$}(output);
    \draw [-] (output) |- node {} (system1);
    \draw [-] (system1) -| node {$U_{I}$} (input);
    %\draw [->] (y) |- (sum2);
    %\draw [->] (a1) -- (controller);
    % \draw [->] (y1) -- ++(0,-3cm) -| node[pos=0.99] {$-$} 
    %     node [near end] {$\psi$} (sum1);
        %  \draw [->] (y) -- ++(0,-2cm) -| node[pos=0.99] {$-$} 
        % node [near end] {$r$} (sum2);
\end{tikzpicture}
    \caption{Architecture of integrated guidance and control for MAV}
    \label{igc1}
\end{figure}

{\label{igc}}
\noindent  In separate guidance and control, the input to the actuator is computed in two  sequential stages from the navigation data. Whereas in integrated guidance and control, the input to the actuator is computed in a single step from the navigation data.  The block diagram of integrated guidance and control  (IGC) is given in Fig. \ref{igc1}. The integrated guidance and control block generates the actuator input ($U_I$) based on the output ($Y_I$) and the next waypoint position ($x_n,y_n)$. $U_P$ is the actuator output that deflects the elevator, rudder control surfaces and 
also changes the thrust input.

    % \begin{figure}[!h]
    %          \begin{center}
    %          \includegraphics[trim = 0cm 0cm 0cm 0cm, clip=true, scale=0.4]{IGCbasblk.PNG}
    %          \caption{Block diagram for integrated guidance and control}
    %          \label{igc1}
    %          \end{center}
    %     \end{figure}
The measured output $Y_I$ is given by
\begin{eqnarray}
Y_I=[q,\theta,h, p, r, \phi, d]^T 
\label{yigc1}
\end{eqnarray}
The yaw angle $\psi$ , the current position of the MAV ($x,y$) and the next waypoint position ($x_f,y_f$) is used for the computation of the miss distance $d$  as given in (\ref{sigma1}) to (\ref{Glawd}). Here the heading angle $\chi \approx \psi$ assuming the angle of side slip ($\beta$) to be small.  
% Since the yaw angle $\psi$  is used in computing the miss distance d, it is not included as a part of the measured output $Y_I$ and in the state vector in the linear state space model for the IGC.
Let $x_{1}=d$, and $\dot{x_{1}}=\dot{d}=x_{2}$. The state variables used in the linear  state space model of integrated guidance and control logic are 
\begin{eqnarray}
 X_I=[X,X_g,X_a]^T
\label{state12}
 \end{eqnarray}
where $X$ is given in (\ref{Lintable1}),
% \begin{eqnarray}
%  X=[\widetilde{ u }, \widetilde{ w }, \widetilde{ q }, \widetilde{ \theta }, \widetilde{ h }, \widetilde{ v }, \widetilde{ p } ,\widetilde{ r },\widetilde{ \phi }]^T
% \label{state11}
%  \end{eqnarray}
$X_g=[\widetilde{ x_{1}}, \widetilde{ x_{2}}]$ are the states corresponding to the guidance logic and $X_a=[{\widetilde{ \delta}}_{ep},{\widetilde{ \delta}}_{ep1},{\widetilde{ \delta}}_{er},{\widetilde{ \delta}}_{er1}]$ corresponds to second order elevator and rudder servo actuator transfer function. Thus
\begin{eqnarray}
 X_I=[\widetilde{ u }, \widetilde{ w }, \widetilde{ q }, \widetilde{ \theta }, \widetilde{ h }, \widetilde{ v }, \widetilde{ p } ,\widetilde{ r },\widetilde{ \phi }, \widetilde{ x_{1}}, \widetilde{ x_{2}}, {\widetilde{ \delta}}_{ep},{\widetilde{ \delta}}_{ep1},{\widetilde{ \delta}}_{er},{\widetilde{ \delta}}_{er1} ]^T
\label{state13}
 \end{eqnarray}
In (\ref{state13}), $\widetilde{ x_{1}}, \widetilde{ x_{2}}$ are the  variables obtained through the linearization of $x_1$ and $x_2$. The first step in the linearization process is to find a relation between  $\dot{x_2}$($\ddot{d}$) and the acceleration command $a_c$. The second step is to find relation between $a_c$ and the turn rate of the vehicle which is a function of $\dot{\psi}$ and $\dot{\beta}$. These two steps will lead to the integration of $X_g=[\widetilde{ x_{1}}, \widetilde{ x_{2}}]$ to the state vector $X$ to form $X_I$, since $\dot{\psi}$ and $\dot{\beta}$ is a function of $X$ . As a first step, the commanded acceleration  is simplified and  is valid for small angles of $\rho$ as  given in the below theorem.\\
\textit{Theorem I}:
For small angle $\rho$, the commanded acceleration can be written as
\begin{equation}
    a_{c}= \frac{1}{N-1}\ddot{d}
    \label{th1}
\end{equation}
\textit{Proof}: The approximations  $sin\rho=\rho$ and $cos\rho=1$  are done for  small angle $\rho$. From (\ref{Glawd}),
\begin{eqnarray}
d=r_{a}\rho
\label{sm1}
 \end{eqnarray}
Differentiating the above equation gives the following,
\begin{eqnarray}
\dot{d}=\dot{r_{a}}\rho+r_{a}\dot{\rho}
\label{sm2}
 \end{eqnarray}
From (\ref{ra1}),
% \begin{eqnarray}
% \dot{r_{a}}=\frac{d}{dt}(\sqrt{(x_{f}-x_{0})^{2}+(y_{f}-y_{0})^{2}})
% \label{sm4}
%  \end{eqnarray}
% which can be calculated as
\begin{eqnarray}
\dot{r_{a}}=-V_{a}
\label{sm4}
\end{eqnarray}
% and for  the small angle approximation,
% \begin{eqnarray}
% \dot{r_{a}}=-V_{a}
% \label{sm5}
%\end{eqnarray}
From (\ref{rho1}),
\begin{eqnarray}
\dot{\rho}=-\dot{\chi}-\dot{\sigma}
\label{sm6}
 \end{eqnarray}
\noindent Using the relation $a_c=V_a \dot{\chi}$ and from (\ref{gdlawf}) 
\begin{eqnarray}
\dot{\chi}=\frac {NV_{a}d}{{r_{a}}^2}
\label{sm7}
\end{eqnarray}
From (\ref{losr}) and (\ref{Glawd}) we obtain,  
\begin{eqnarray}
\dot{\sigma}=-\frac{V_{a}d}{{r_{a}}^2}
\label{sm9}
 \end{eqnarray}
Substituting (\ref{sm7}) and (\ref{sm9}) into (\ref{sm6})  gives,
\begin{eqnarray}
\dot{\rho}=-(N-1)\frac{V_{a}d}{{r_{a}}^2}
\label{sm99}
 \end{eqnarray}
%using equation \ref{Glawd},  $\dot{\sigma}$ can be simplified as
%\begin{eqnarray}
%\dot{\sigma}=\frac{V_{a}d}{{r_{a}}^2}
%\label{sm9}
% \end{eqnarray}
From (\ref{sm4}), (\ref{sm99})  and (\ref{sm2})  we obtain,
\begin{eqnarray}
\dot{d}=-\frac{NV_{a}d}{r_{a}}
\label{sm10}
 \end{eqnarray}
Differentiating  the above equation and using (\ref{Glawd})   gives,
% \begin{eqnarray}
% \ddot{d}=-NV_{a}\frac{d}{dt}(\frac{d}{r_{a}})
% \label{sm11}
%  \end{eqnarray}
% From equation \ref{Glawd} and small angle approximation, 
\begin{eqnarray}
\ddot{d}=-NV_{a}\frac{d}{dt}(\rho)
\label{sm12}
 \end{eqnarray}
Using (\ref{sm7}) and (\ref{sm9}),  (\ref{sm12}) can be rewritten as
\begin{eqnarray}
\ddot{d}=(N-1)\frac{{{NV_{a}}^2}d}{{r_{a}}^2}
\label{sm13}
 \end{eqnarray}
Noting that $a_{c}=\frac{N{V_{a}}^2d}{{r_{a}}^2}$,  the desired result is obtained.

\begin{figure}[!h]
                 \begin{center}
                  \includegraphics[trim = 0cm 0cm 0cm 0cm, clip=true, scale=0.50]{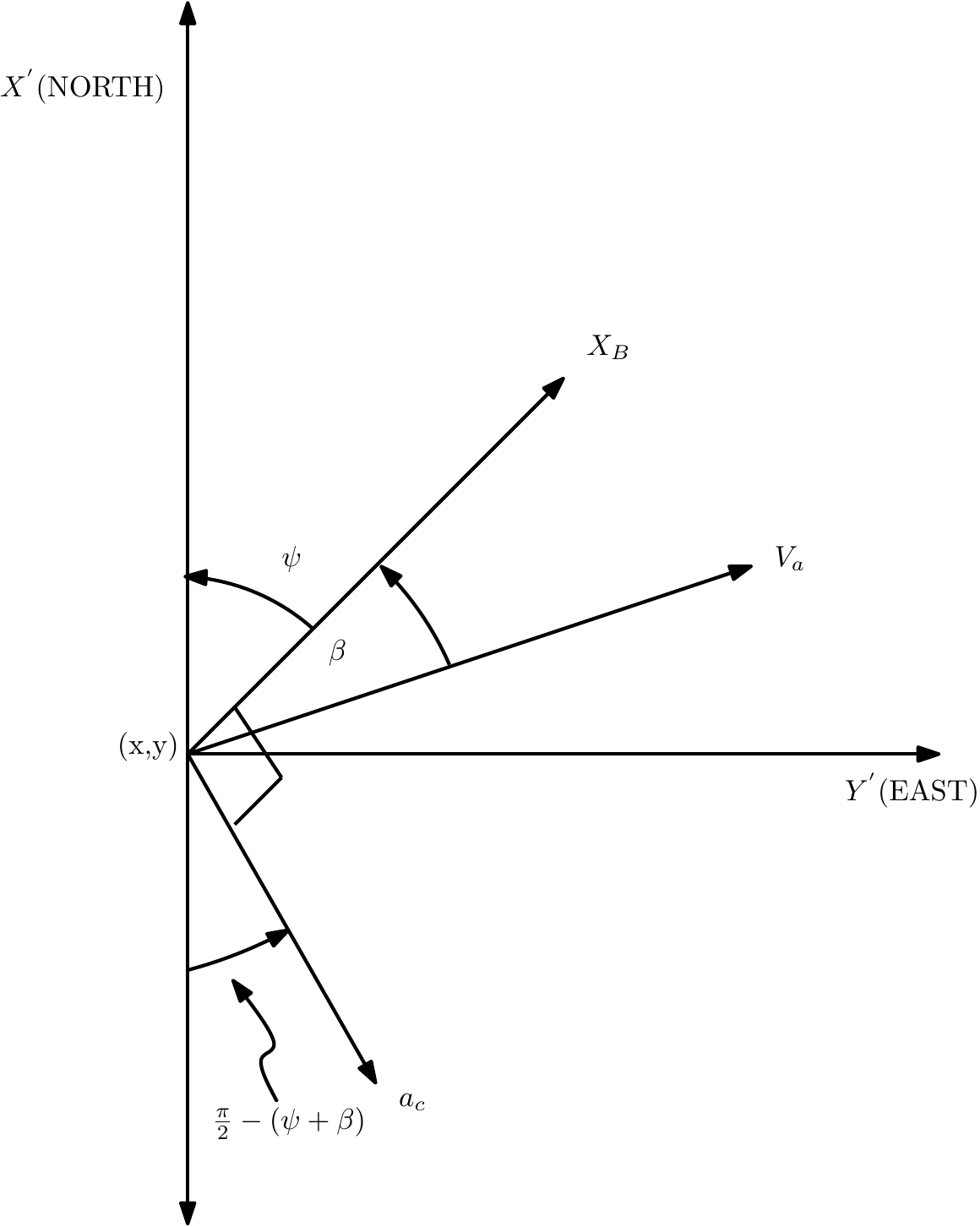}
                 \caption{ Body fixed lateral inertial plane of the MAV}
                 \label{3D77}
                 \end{center}
                 \end{figure}

 \noindent The  approximation of the non-linear guidance law  that is valid for small angles of $\rho$  is given in  (\ref{th1}). Now as the second step to enable  the linearization process, a relationship between $a_c$ and  $\dot{\psi}$ , $\dot{\beta}$ is derived.  In Fig. \ref{3D77}, $X'Y'$ is the body fixed lateral inertial plane of the MAV with X' pointing towards north and Y' pointing towards east, $X_{B}$ denotes the body X axis, $\beta$ is the sideslip angle, $\psi$ is the yaw angle. The origin of the body fixed lateral inertial plane is the current position of the MAV denoted by $(x, y)$. When an acceleration $a_c$ is applied perpendicular to the velocity vector in the yaw plane as shown in the Fig. \ref{3D77}, the  MAV will move in a circular path  in the body fixed lateral inertial plane with an angular velocity  $\dot{\psi}+\dot{\beta}$ . Then the applied acceleration can be equated to $V_{a}(\dot{\psi}+\dot{\beta})$ as given in  (\ref{acc12}).
The heading angle $\chi=\psi+\beta$. From the Fig. \ref{3D77}, we obtain
\begin{eqnarray}
a_{c}=V_{a}(\dot{\psi}+\dot{\beta})
\label{acc12}
\end{eqnarray}
The expression for $\dot{\psi}$ and $\dot{\beta}$  is given in (\ref{psieq}) and (\ref{betaeq}) respectively.
\begin{eqnarray}
\dot{\psi}=qsin\phi sec\theta +rcos\phi sec\theta 
\label{psieq}
\end{eqnarray}
\begin{eqnarray}
\dot{\beta}=\frac{1}{V_a}(pw-ru+gcos\theta sin\phi+\frac{Y_a}{m} )
\label{betaeq}
\end{eqnarray}
where $Y_a$ is the side-force given by
\begin{eqnarray}
Y_{a}=\bar{Q}SC_y
\label{sforce1}
\end{eqnarray}
where $\bar{Q}=0.5 \rho_a {V_{a}}^2$ is the dynamic pressure in $N/m^2$ and $C_y$ is the side-force coefficient given by
\begin{eqnarray}
C_{y}=(C_{y\beta}\beta+C_{yp}(\frac{0.5b}{V_{a}})p+C_{yr}(\frac{0.5b}{V_{a}})r+C_{y\delta r} \delta_{r})
\label{sforce2}
\end{eqnarray}
where $C_{y\beta}$, $C_{yp}$, $C_{yr}$ and $C_{y\delta_r}$ denotes the change in side-force coefficient due to $\beta$, $p$, $r$ and $\delta_r$ respectively.
% Using  (\ref{psieq}) and (\ref{betaeq}),  (\ref{acc12}) becomes
% \begin{eqnarray}
% a_{c}=V_{a}(qsin\phi sec\theta +rcos\phi sec\theta)+(pw-ru+gcos\theta sin\phi+\frac{Y_a}{m} ) 
% \label{acc13}
% \end{eqnarray}
From (\ref{acc12}), (\ref{psieq}) and (\ref{betaeq}) we can see that the applied acceleration $a_c$ is a function of the state variables and the rudder input.
%Let $x_{3}=d$, and $\dot{x_{3}}=\dot{d}=x_{4}$. 
%  Then from (\ref{sm15}) and  (\ref{acc13}),
% \begin{eqnarray}
% \dot{x_{4}}=(N-1)a_{c}
% \label{xcvb}
% \end{eqnarray}
% \begin{eqnarray}
% \dot{x_{4}}=(N-1)(V_{a}(qsin\phi sec\theta +rcos\phi sec\theta)+(pw-ru+gcos\theta sin\phi+\frac{Y_a}{m} ) )
% \end{eqnarray}
Thus the state equations for the guidance logic are
\begin{eqnarray}
\dot{x_{1}}=x_{2}
\label{x31}
\end{eqnarray}
\begin{eqnarray}
\dot{x_{2}}=(N-1)a_c
\label{x41}
\end{eqnarray}
  Equations  (\ref{x31}) and (\ref{x41}) are linearized and augmented with the MAV dynamic model to obtain the state space model for integrated guidance and control. Let the linearized variables for  $x_{1}$ and $x_{2}$ be  $\widetilde{ x_{1}}$ and $\widetilde{ x_{2}}$ respectively. Then the linear state equation and output equation for the integrated guidance and control for two dimensional space is given by
 \begin{eqnarray}
\dot {X_I}=AX_I+BU_I \\
Y_I=CX_I
\end{eqnarray}
% where $X_I=[\widetilde{ u }, \widetilde{ w }, \widetilde{ q }, \widetilde{ \theta }, \widetilde{ h }, \widetilde{ v }, \widetilde{ p } ,\widetilde{ r },\widetilde{ \phi }, \widetilde{ x_{3}}, \widetilde{ x_{4}}]^T$,  $Y=[\widetilde{ q }, \widetilde{ \theta },\widetilde{ h }, \widetilde{ p }, \widetilde{ r }, \widetilde{ \phi }, \widetilde{ x_{3}}]^T$ and  $U_I=[\widetilde{ \delta_{ep}},\widetilde{ \delta_{rp}},\widetilde{ \delta_{thp}}]^T$. The derivative of the miss distance, ${\widetilde{ x_{4}}}$ is not taken as a measured variable. From (\ref{sm10}) we can see that the computation of  ${\widetilde{ x_{4}}}$ requires the knowledge of the airspeed $V_a$. Reliable air speed information is not available due to the lack of low weight good quality sensors and heavy propeller wake disturbances that corrupts the measurement.

% The second order model for elevator servo and rudder servo is given in equations \ref{servo1} and \ref{servo2} respectively.
%\begin{eqnarray}
%\dot{\widetilde{ \delta}}_{ep}={\widetilde{ \delta}}_{ep1}\\
%\dot{\widetilde{ \delta}}_{ep1}=-2367{\widetilde{ \delta}}_{ep}-72.22{\widetilde{ \delta}}_{ep1}+2367\widetilde{ \delta_{e}}
%\label{servo1}
%\end{eqnarray}
%\begin{eqnarray}
%\dot{\widetilde{ \delta}}_{er}={\widetilde{ \delta}}_{er1}\\
%\dot{\widetilde{ \delta}}_{er1}=-2367{\widetilde{ \delta}}_{er}-72.22{\widetilde{ \delta}}_{er1}+2367\widetilde{ \delta_{r}}
%\label{servo2}
%\end{eqnarray}

\noindent where $X_I$ is given in (\ref{state13}) and
% \begin{equation}
%   X_I=[\widetilde{ u }, \widetilde{ w }, \widetilde{ q }, \widetilde{ \theta }, \widetilde{ h }, \widetilde{ v }, \widetilde{ p } ,\widetilde{ r },\widetilde{ \phi }, \widetilde{ x_{3}}, \widetilde{ x_{4}},{\widetilde{ \delta}}_{ep},{\widetilde{ \delta}}_{ep1},{\widetilde{ \delta}}_{er},{\widetilde{ \delta}}_{er1} ]^T   
% \end{equation}
\begin{equation}
U_I=[\widetilde{ \delta_{e}},\widetilde{ \delta_{r}},\widetilde{ \delta_{t}}]^T    
\end{equation}
\begin{equation}
 Y_I=[\widetilde{ q }, \widetilde{ \theta },\widetilde{ h }, \widetilde{ p }, \widetilde{ r }, \widetilde{ \phi }, \widetilde{ x_{1}}]^T   
\end{equation}

\subsection{Controller design for IGC framework} 
 
\noindent For the trim conditions at $V_a=8m/s$, $\dot{\psi}=0.8 rad/s$ and climb rate $\dot{h}=1 m/s$, the poles of  linear state space model corresponding to IGC framework is given in Table  \ref{table:igcopen1}. The spiral mode is unstable for the open loop system.
\begin{table}[!ht]
\caption{Poles of linear state space model - IGC framework}
\centering
\begin{tabular}{|c|c|}
\hline
Short period mode ($\omega_{sp}$, $\zeta_{sp}$) & 35.6 $rad/s$, 0.285 \\
\hline
Phugoid mode ($\omega_{ph}$, $\zeta_{ph}$) & 1.15 $rad/s$, 0.219 \\
\hline
Dutch roll mode, ($\omega_{dr}$, $\zeta_{dr}$) & 43.8 $rad/s$, 0.337\\
\hline
Roll subsidence mode  & -2.44\\
\hline
Spiral mode  & 1.51 (unstable) \\
\hline
Elevator servo ($\omega_{e}$, $\zeta_{e}$) & 48.7 $rad/s$, 0.742 \\
\hline
Rudder servo ($\omega_{r}$, $\zeta_{r}$) & 48.7 $rad/s$, 0.742 \\
\hline
Poles for $\widetilde{ h }$, $ \widetilde{ x_{1}}, \widetilde{ x_{2}}$ & 0, 0, 0 \\
\hline
\end{tabular} \\
\label{table:igcopen1}
\end{table}

\begin{figure}
    \centering
   \begin{tikzpicture}[auto,>=latex']
    % We start by placing the blocks
    \node [input, name=input] {};
    \node [sum, right = of input] (sum) {};
     \node [output, right =of sum, node distance=1cm] (a2) {};
    \node [block, right = of a2, node distance=1cm] (controller) {$K_c$};
     %\node [block, right = of controller,
      %      node distance=3cm] (act) {$G_a$};
       \node [output, right =of controller, node distance=1cm] (act) {};
    \node [block, right = of act,
            node distance=2cm] (system) {MAV};
    % We draw an edge between the controller and system block to 
    % calculate the coordinate u. We need it to place the measurement block. 
   
    \draw [->] (controller) -- node[name=u] {$U_I$} (act);
    
    \draw [->] (act) -- node[name=u] {} (system);
    \node [output, right =of system,node distance=2cm] (output) {};
   % \node [, below =of controller] (a1) {$Y_2$};
    \node [block, above =of controller] (w1) {$W_1$};
    \node [block, below =of system] (w3) {$W_3$};
    \node [block, right = of output,
            node distance=2cm] (w2) {$W_2$};
            
             \node [output, right =of w2,node distance=2cm] (w2o) {};
             
              \node [output, right =of w1,node distance=2cm] (w1o) {};
              
               \node [output, right =of w3,node distance=2cm] (w3o) {};
           \draw [draw,->] (w2) -- node {$Z_2$} (w2o);
           
           \draw [draw,->] (w1) -- node {$Z_1$} (w1o);
           
           \draw [draw,->] (w3) -- node {$Z_3$} (w3o);

    % Once the nodes are placed, connecting them is easy. 
    \draw [draw,->] (input) -- node {$Y_r$} (sum);
    \draw [->] (sum) -- node {$E$} (a2);
    \draw [->] (a2) -- node {} (controller);
     \draw [->] (a2) |- node {} (w1);
     
     \draw [->] (act) |- node {} (w3);
    
    \draw [->] (system) -- node [name=y] {$Y_I$}(output);
    \draw [->] (output) -- node [name=y] {}(w2);
   % \draw [->] (y) -- (controller);
    %\draw [->] (a1) -- (controller);
    \draw [->] (y) -- ++(0,-4cm) -| node[pos=0.99] {$-$} 
        node [near end] {$Y_1$} (sum);
         \draw [->] (y) -- ++(0,-3.2cm) -| node[pos=0.99] {} 
        node [near end] {$Y_2$} (controller);
\end{tikzpicture}
    \caption{$H_{\infty}$ control architecture for IGC framework}
    \label{mav1con}
\end{figure}
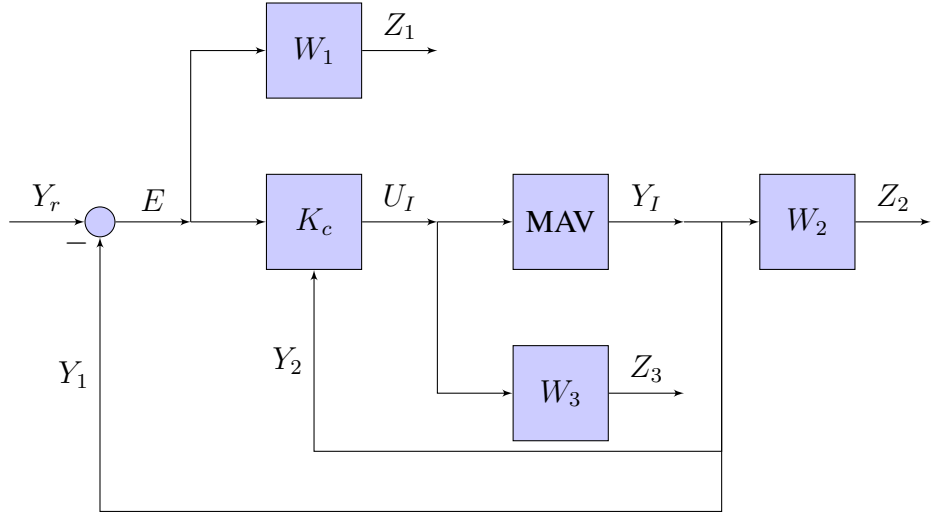

\noindent The controller is synthesized in $H_{\infty}$ robust control framework as it can handle uncertainties associated with the MAV dynamics. The architecture of $H_{\infty}$ control for IGC framework is shown in Fig. \ref{mav1con}. Here the reference signal $Y_r=[\widetilde{h}_r,\,\widetilde{ x_{1}}_r]^T$, and the measured outputs $Y_1=[\widetilde{h},\,\widetilde{ x_{1}}]^T$, $Y_2=[\widetilde{ q }, \widetilde{ \theta }, \widetilde{ p }, \widetilde{ r }, \widetilde{ \phi }]^T$. The weighting transfer functions $W_1$, $W_2$ and $W_3$ are low pass filter, high pass filter and constant respectively \cite{jiankun}. $Z_1$, $Z_2$, $Z_3$ are the performance outputs to be minimized.

\noindent The controller synthesis is performed in the discrete time domain using the algorithm described in \cite{hari2}. The controller sampling time is 0.02 seconds (the standard for servo control). The algorithm is summarized as given below.  The discrete time equivalent linear state space model of the generalized plant shown in Fig. \ref{mav1con} is given in  (\ref{cplant1}) to (\ref{cplant5}).
\begin{eqnarray}
{X_{dg}}(k+1)=A_{dg}X_{dg}(k)+B_{du}{U_{dc}}(k)+B_{dw}{W_{dd}}(k)
\label{cplant1}
\end{eqnarray}
 \begin{eqnarray}
{Z_{d1}}(k)=C_{d1}X_{dg}(k)+D_{d11}{U_{dc}}(k)+D_{d12}{W_{dd}}(k)
\label{cplant2}
\end{eqnarray}
 \begin{eqnarray}
{Z_{d2}}(k)=C_{d2}X_{dg}(k)+D_{d21}{U_{dc}}(k)+D_{d22}{W_{dd}}(k)
\label{cplant3}
\end{eqnarray}
 \begin{eqnarray}
{Z_{d3}}(k)=C_{d3}X_{dg}(k)+D_{d31}{U_{dc}}(k)+D_{d32}{W_{dd}}(k)
\label{cplant4}
\end{eqnarray}
 \begin{eqnarray}
Y_d(k)=C_dX_{dg}(k)
\label{cplant5}
\end{eqnarray} 
\noindent The controller design objective is to synthesis a SOF control law,
\begin{equation}
{U_{dc}}(k)=F_dY_d(k)
\label{igclaw}
\end{equation}
that minimizes the weighted sensitivity, weighted complimentary sensitivity and weighted control input sensitivity function. The gain matrix $F_d$ is obtained by solving the LMI (linear matrix inequality) given in  (\ref{lmi1}). The matrix $N_d$ is selected using the genetic algorithm (GA) for minimizing a performance index considering the closed loop damping ratio and stability of spiral mode and the resultant LMI is solved for $P_d$ and $F_d$. The matrix $P_d>0$ for a feasible solution $F_d$ that stabilizes the closed loop plant.
\begin{eqnarray}
 \left(
\begin{array}{cc}
- P_d & -(A_{dg}+B_{du}F_dC_d)^{T}N_d^{T}  \\
  -N_d(A_{dg}+B_{du}F_dC_d)   &    P_d-N_d-N_d^{T}
\end{array}
 \right)<0
\label{lmi1}
\end{eqnarray}

\noindent The controller gain matrix $F_d$ is given in  (\ref{kgain2D}). 
% The gain in the last column corresponding to the variable $ \widetilde{ x_{3}}$ is smaller when compared to other feedback gains. This is because the output $ \widetilde{ x_{3}}$ is the miss distance in meters and its magnitude is much higher than the other outputs. 
%\begin{equation}
%F= \left[ \begin {array}{ccccccc} - 0.001997&- 0.03378&- 0.001057& 0.0001952& 0.0& 0.0&- 0.1008\\\noalign{\medskip} 0.0& 0.1357& 0.006784&- 0.01661& 0.0007563&- 0.02343&- 0.0007457\\\noalign{\medskip}- 1.152& 41.69& 0.006974&- 4.841& 0.0& 0.0& 0.0\end {array} \right] 
%\label{kgain2D}
%\end{equation}
\begin{equation}
F_d= [F_{d1},\,F_{d2}]
\label{kgain2D}
\end{equation}
\begin{equation}
F_{d1}= \left[ \begin {array}{ccc} -0.0060&- 0.3378&- 0.8032\\\noalign{\medskip} 0& 0.0987& 0\\\noalign{\medskip}- 1.152& 36.54& 0.0042\end {array} \right] 
\label{kgain2D1}
\end{equation}
\begin{equation}
F_{d2}= \left[ \begin {array}{cccc}  0.0002& -0.0007& 0&- 0.1008\\\noalign{\medskip}  -0.0528&0.0008&- 0.0937&-0.0002\\\noalign{\medskip}- 5.325& -19.1& 0.0004& 0.0091\end {array} \right] 
\label{kgain2D2}
\end{equation}
\noindent For the trim conditions at $V_a=8m/s$, $\dot{\psi}=0.8 rad/s$ and climb rate $\dot{h}=1 m/s$, the dynamic modes of IGC closed loop linear state space model in continuous time domain is given in Table  \ref{table:igcclose1}. From the Table  \ref{table:igcclose1}, we can see that the spiral mode is stabilized and the damping ratio of other modes have improved when compared to open loop plant. The three poles at the origin corresponding to $\widetilde{ h }$, $ \widetilde{ x_{1}}$ and $\widetilde{ x_{2}}$ are placed in the left half of the s-plane.

\begin{table}[!ht]
\caption{Poles of closed loop IGC linear state space model}
\centering
\begin{tabular}{|c|c|}
\hline
Short period mode ($\omega_{sp}$, $\zeta_{sp}$) & 36.8 $rad/s$, 0.404 \\
\hline
Phugoid mode ($\omega_{ph}$, $\zeta_{ph}$) & 2.25 $rad/s$, 0.323 \\
\hline
Dutch roll mode, ($\omega_{dr}$, $\zeta_{dr}$) & 30.0 $rad/s$, 0.429\\
\hline
Roll subsidence mode  & -15.5\\
\hline
Spiral mode  & -3.24 \\
\hline
Pole for $\widetilde{ h }$ & -1.50 \\
\hline
Poles for $ \widetilde{ x_{1}}$, $\widetilde{ x_{2}}$ ($\omega$, $\zeta$) & 0.1 $rad/s$, 0.343 \\
\hline
Elevator servo ($\omega_{e}$, $\zeta_{e}$) &44.4 $rad/s$, 0.671 \\
\hline
Rudder servo ($\omega_{r}$, $\zeta_{r}$) &49.5 $rad/s$, 0.603 \\
\hline
\end{tabular} \\
\label{table:igcclose1}
\end{table}
%A plot of sensitivity (S) and complementary sensitivity (T) is given in figure \ref{sandt2d}. From figure \ref{sandt2d}, the closed loop system has good tracking and disturbance rejection performance till 4 rad/s. A low value of -20 dB is achieved for T beyond 10 rad/s. This indicates rejection of  sensor noise and robustness against unmodeled dynamics after a frequency of 10 rad/s. A rise in the value of T between 30  rad/s and 40 rad/s is due to the closed loop natural frequency of short period and Dutch roll mode appearing in that range.
% \begin{figure}[!ht]
%                 \begin{center}
%                 \includegraphics[trim = 0cm 0cm 0cm 0cm, clip=true, scale=0.4]{SandTfor2D.PNG}
%                 \caption{S and T plot for the closed loop IGC plant for 2D waypoint navigation}
%                 \label{sandt2d}
%                 \end{center}
%                 \end{figure}
                 
\subsection{ Waypoint navigation algorithm using IGC framework}
%The minimum error that can be achieved in  following a waypoint  depends upon the accuracy of the GPS used. From Table \ref{}, the circular error probability (CEP) of the GPS used for waypoint navigation is 2.5 m. A CEP of 2.5 m indicates for the 50$\%$ of the time the error in position is less than 2.5 m and for 43$\%$ time, the error in position is between 2.5 m to 5 m and for the rest 7$\%$ of the time the error in position is between 5 m to 7.5 m. Considering the worst case GPS error, an error of 10 m is set for the waypoint navigation in the absence of wind disturbances.
\noindent Inputs to the algorithm are the current $(x,y)$ position of the MAV obtained from GPS, the desired waypoint position $(x_f,y_f)$, altitude $h$ obtained from altimeter, rate gyro outputs p, q, r and Euler angles $\phi$, $\theta$, $\psi$. The steps are summarized in \textit{Algorithm} I. For a given waypoint, the angle $\rho$ is computed from (\ref{rho1}) and a constant roll angle command is issued if $|\rho|>20^0$. Here the small angle approximation $sin\rho=\rho$ and $cos\rho=1$ is used for $|\rho| \leq 20^0$ and the expression given in (\ref{th1}) is valid for the same. The roll angle control for 150 mm MAV given in \cite{hari1} is used here. The altitude is maintained constant while executing the roll angle command. The MAV follows a constant roll angle command $\phi_{ref}=sgn(\rho)\phi_r$ for $|\rho|>20^0$. Here  $\phi_r$ is a constant value less than the maximum allowed roll angle $\phi_{max}$ of the MAV. When $|\rho| \leq 20^0$, the feasibility of reaching the waypoint is verified using (\ref{rcons1}) and subsequently  the control law  given in (\ref{igclaw}) is employed as the IGC framework is valid. When the range ($r_a$) is less than a pre-defined threshold $r_t$, the waypoint is considered to be reached and the entire process is repeated for a new waypoint. \\

%\begin{algorithm}
%\hrule
%\vspace{0.1cm}
%\textit{Algorithm} I: Waypoint navigation using IGC framework
%\hrule
%\SetAlgoLined
%\vspace{0.1cm}
%STEP I:  {Compute $\rho$ from (\ref{rho1}) }.\\
%STEP II:  \\ \eIf{$|\rho|>20^0$}{
%  Issue a  roll angle hold command $\phi_{ref}=sgn(\rho)\phi_{r}$, where $0<\phi_{r}<\phi_{max} $ and \textbf{goto} STEP I.\\
%  
% }
% {\eIf{$r_a>2 R_{min} sin\rho$}{ 
%   Use the controller structure given in (\ref{igclaw}) and \textbf{goto} STEP III.\\
%   }{
%   Current waypoint is not feasible, \textbf{goto} STEP IV. \\
%  }}
%  STEP III: \\
%  \eIf{$r_a<r_t$}{Waypoint reached, \textbf{goto} STEP IV. }{\textbf{goto} STEP I.}
%  STEP IV: Switch to a new waypoint and \textbf{goto} STEP I.
%  \hrule   
%\end{algorithm}

\textit{Algorithm I}:\\ \hrule
\vspace{0.2 cm}
\noindent STEP I:  Compute $\rho$ from (\ref{rho1}).\\
STEP II: \textbf{else-if}    $|\rho|>20^0$
 Issue a  roll angle hold command $\phi_{ref}=sgn(\rho)\phi_{r}$, where $0<\phi_{r}<\phi_{max} $ and \textbf{goto} STEP I.\\
\textbf{else-if} $r_a>2 R_{min} sin\rho$
   Use the controller structure given in (\ref{igclaw}) and \textbf{goto} STEP III.\\
 Current waypoint is not feasible, \textbf{goto} STEP IV. \\
  STEP III:   \textbf{else-if} $r_a<r_t$ Waypoint reached, \textbf{goto} 
STEP IV \textbf{else} \textbf{goto} STEP I. \\
 STEP IV: Switch to a new waypoint and \textbf{goto} STEP I.
\vspace{0.2 cm}
\hrule

                 \section{Numerical simulation results for waypoint navigation} {\label{numsim}}
\noindent The waypoint navigation algorithm is  simulated for the non-linear six degrees of freedom (6DOF) model of the 150 mm MAV given in \cite{mav1}. Flight test results are presented in \cite{mav1}, validating the non-linear model used for simulation.   Two simulations are performed in which the first one is to demonstrate  straight line path following and  for the second one, the objective is to follow a rectangular path. Constraints are put on the actuators while performing the simulations. The range of elevator deflection is $[-35^o, 15^o]$, rudder deflection is $[-25^o, 25^o]$ and motor thrust is 0.0-0.45 N. The second order transfer function model is used for elevator and rudder servos. The control input update is done at every 0.02 second. The $XY$ position update is done at every 1 second. All other outputs are sampled at a rate of 0.02 second. The waypoint navigation logic is started when the MAV  reach at an altitude of 20 m from the ground level. Altitude feedback is used to maintain the desired altitude of 20 m. For both the simulations, the initial conditions are shown in Table \ref{simtab1}.
\begin{table}[!ht]
          \begin{center}
            \caption{Initial conditions used for the non-linear simulation}
            \begin{tabular}{|p{1.7cm}|p{4.7cm}|}
            \hline
          \textbf{Variable} &  \textbf{Value}   \\
                \hline
           $V_{a}$, $\alpha$, $ \beta$ &  8 m/s, 0.2286  rad, -0.0532 rad  \\
                 \hline
             $\delta_{e}$, $\delta_{r}$, $\delta_t$ &  -0.2592 rad, 0.1255 rad, 0.231 N \\
                  \hline
                     $\phi$, $\theta$, $\psi $&  -0.0361 rad, 0.2304 rad, 0 rad\\
                 \hline
              p, q, r & 0 rad/s, 0 rad/s, 0 rad/s  \\
\hline
u, v, w & 7.78 m/s, -0.43 m/s, 1.81 m/s  \\
 \hline
$x$, $y$, $h$ & 0 m, 0 m, 20 m  \\
                \hline
  \end{tabular}
\label{simtab1}
            \end{center}
            \end{table}

\begin{figure}[!ht]
             \begin{center}
             \includegraphics[height=6cm, width=9cm]{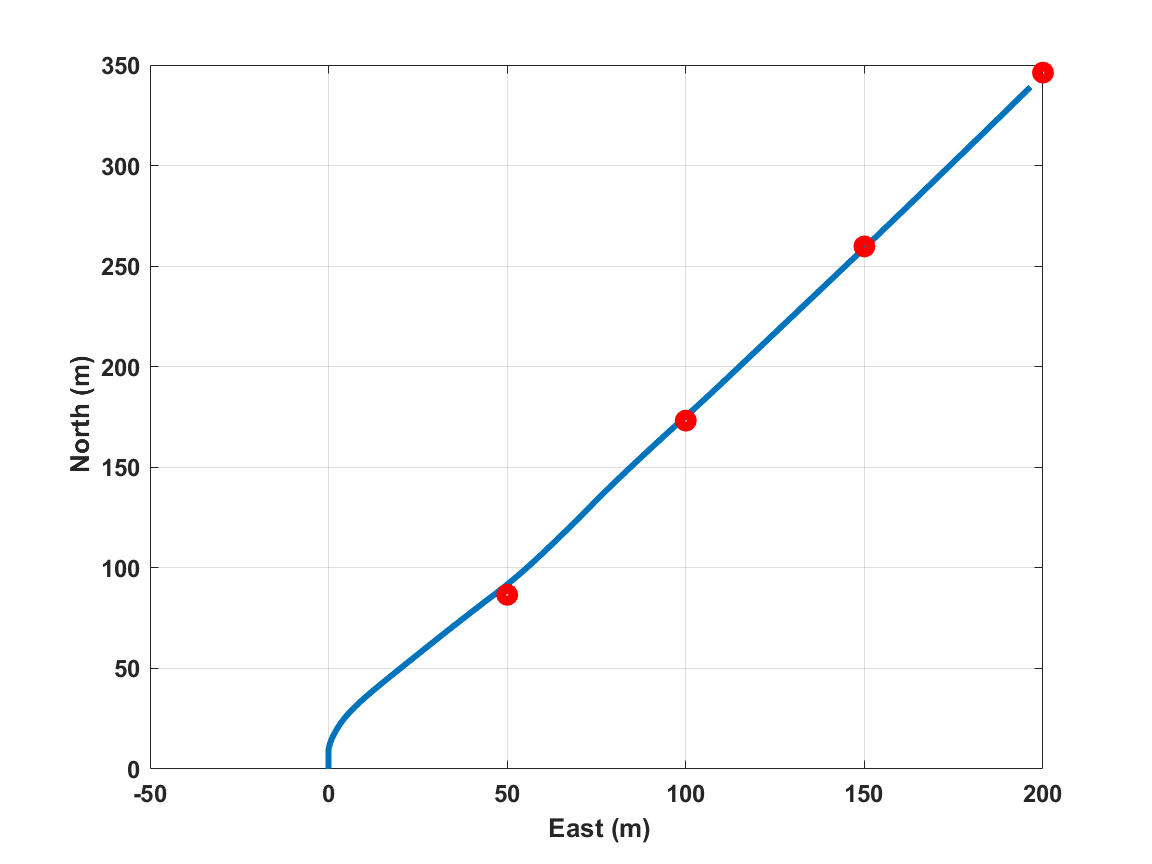}
             \caption{XY plot for MAV  straight line following}
             \label{xyplots1}
             \end{center}
        \end{figure}

\begin{figure}[!ht]
             \begin{center}
             \includegraphics[height=6cm, width=9cm]{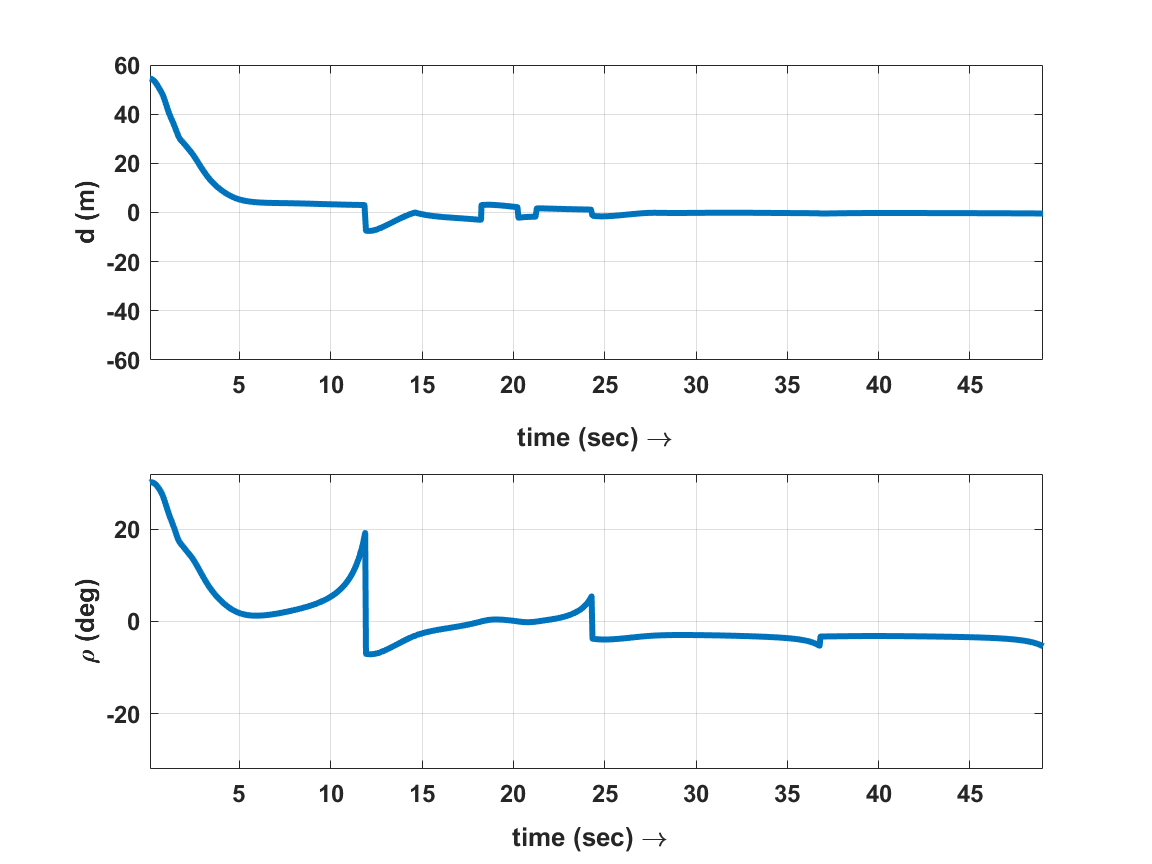}
             \caption{Miss distance and $\rho$ angle plots for MAV  straight line following}
             \label{missds1}
             \end{center}
        \end{figure}

\begin{figure}[!ht]
             \begin{center}
             \includegraphics[height=6cm, width=9cm]{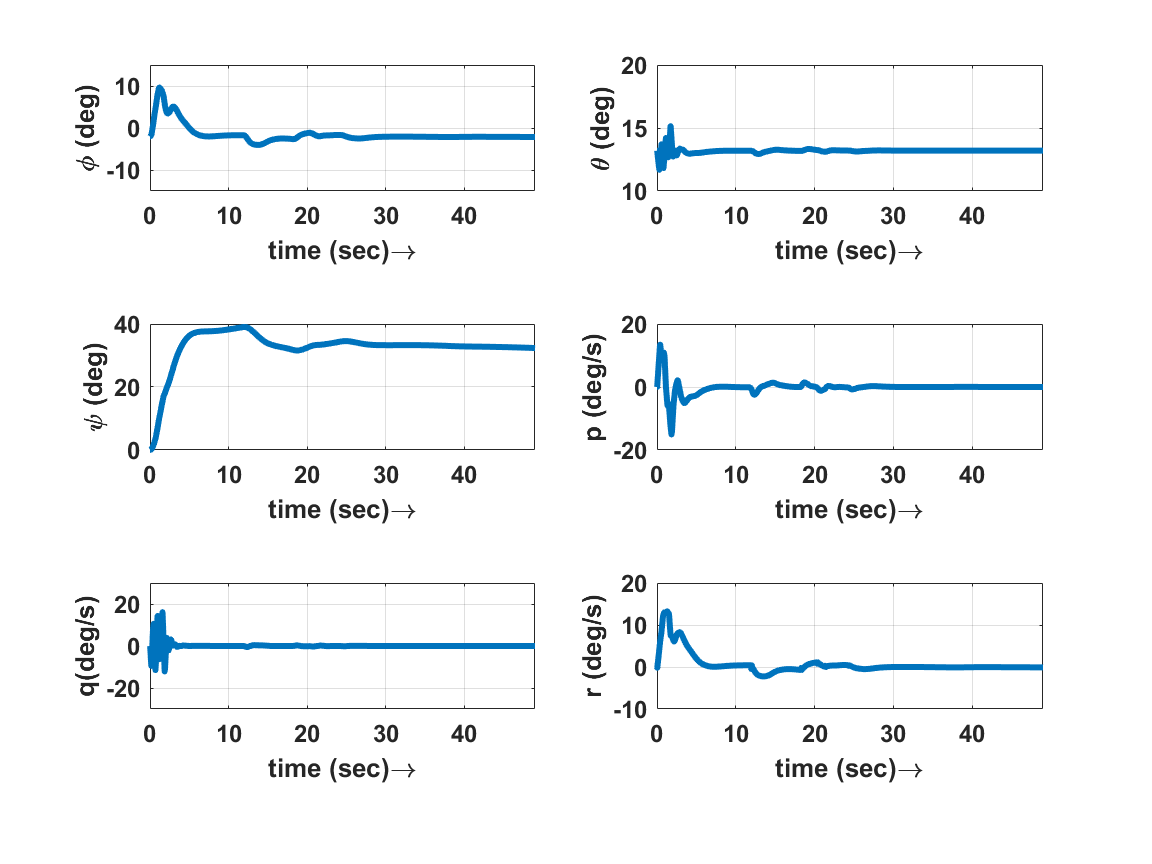}
             \caption{Euler angles and body axis angular rates for MAV  straight line following}
             \label{state1}
             \end{center}
        \end{figure}

\begin{figure}[!ht]
             \begin{center}
             \includegraphics[height=6cm, width=9cm]{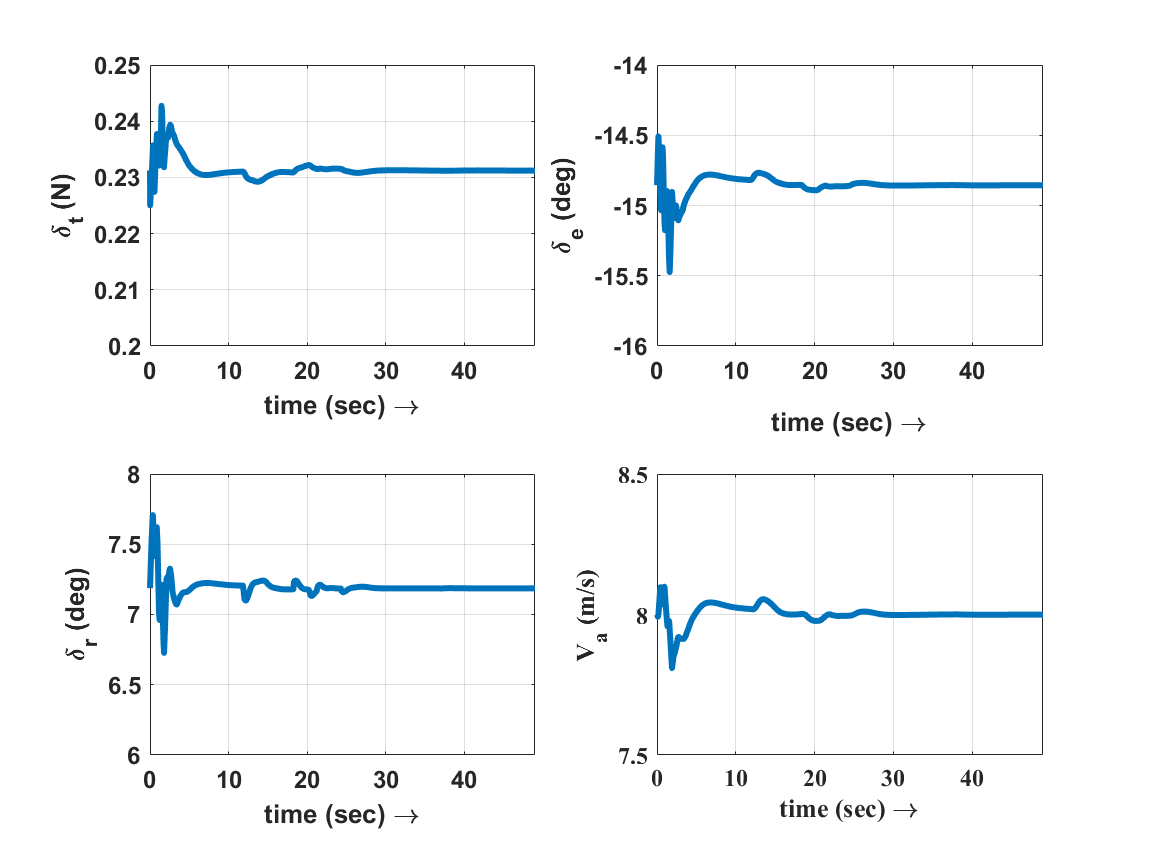}
             \caption{Control inputs and airspeed for  straight line following}
             \label{inpus1}
             \end{center}
        \end{figure}

\noindent The objective of the first simulation is  to follow a straight line.  The XY (North-East) path of the MAV is shown in Fig. \ref{xyplots1}. In Fig. \ref{xyplots1}, red circles represent the waypoints commanded and  the solid blue line represents the path followed by the MAV.  The plot of miss distance and the angle $\rho$ is given in Fig. \ref{missds1}. During the initial stage, the miss distance is high as the first waypoint is not along the line-of-sight (LOS) of the MAV.   After 25 seconds, we can see that the  value of $\rho$ and miss distance is closer to zero as the MAV got aligned to the straight line. 
The small angle approximation used in the derivation of the IGC framework is used for $\rho<20^o$. Initially when  $\rho>20^o$ till 1.4 seconds, a constant roll angle command of $\phi_r=10^o$ is employed. A plot of Euler angles and angular rates are given in Fig. \ref{state1}. The roll angle $\phi$ goes to $10^o$ as shown in Fig. \ref{state1} accordingly.  The plot of control inputs is given in Fig. \ref{inpus1}. The peak amplitude of elevator deflection is less than $-16^o$ and that of the rudder is less than $8^o$. The peak value of elevator deflection is less than $50\%$ of the saturation value of $-35^o$. The peak value of rudder deflection is less than $33\%$ of the saturation value of $25^o$. Thus the control inputs are well within the saturation limits. The variation airspeed is between 7.7 m/s and 8.2 m/s as seen from Fig. \ref{inpus1} and is closer to the trim velocity of 8.0 m/s.

\begin{figure}[!ht]
             \begin{center}
             \includegraphics[height=6cm, width=9cm]{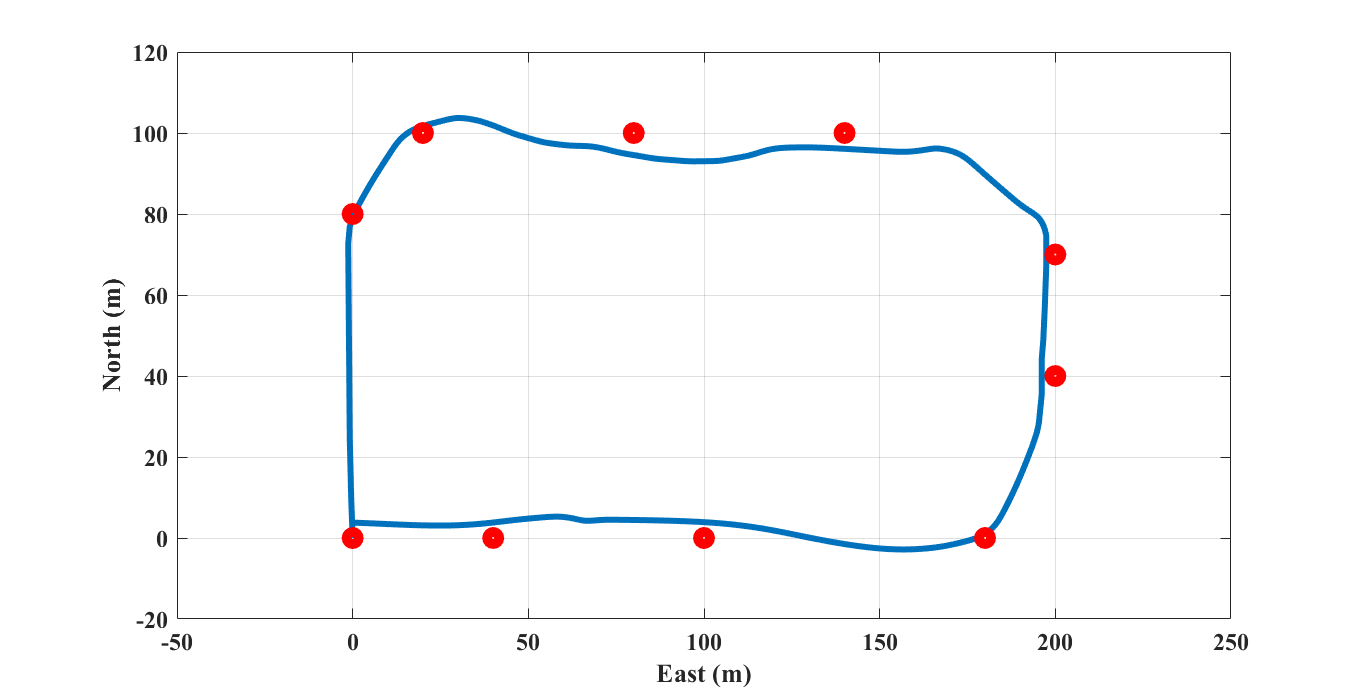}
             \caption{XY plot for rectangular path following}
             \label{xyplots2}
             \end{center}
        \end{figure}

\begin{figure}[!ht]
             \begin{center}
             \includegraphics[height=6cm, width=9cm]{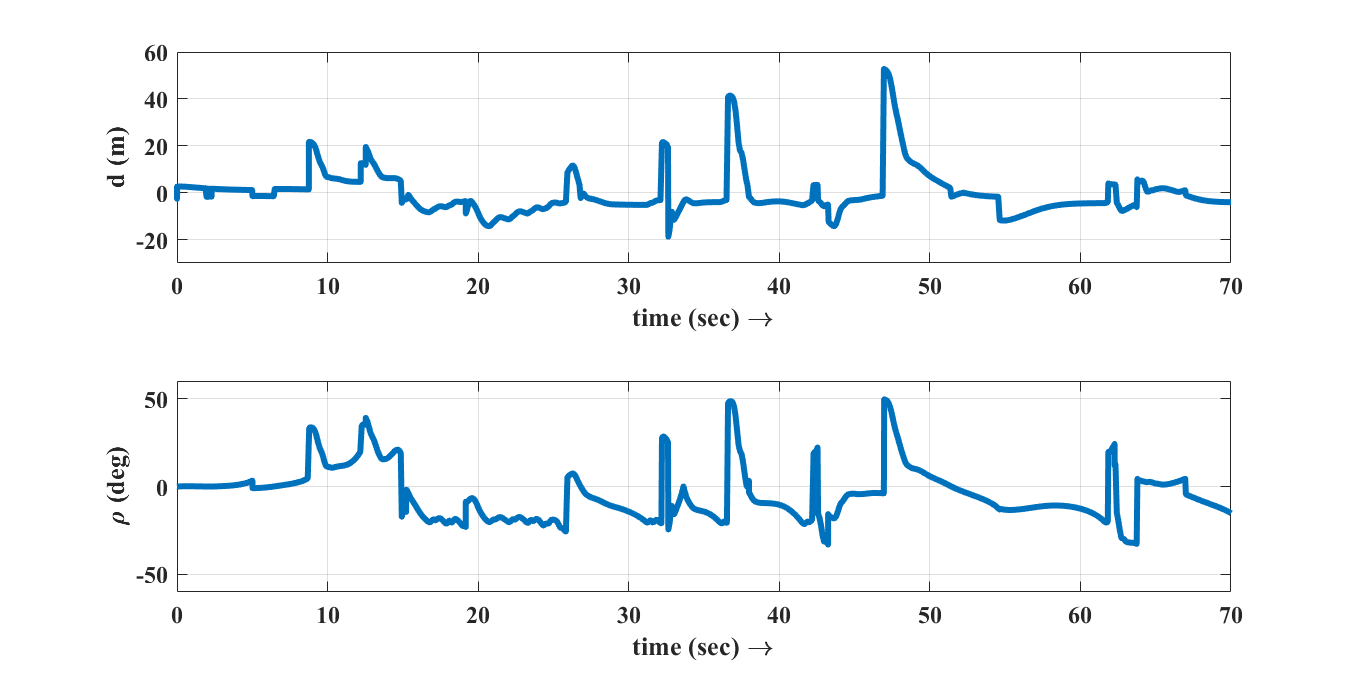}
             \caption{Miss distance and $\rho$ angle plots for rectangular path following}
             \label{missds2}
             \end{center}
        \end{figure}

\begin{figure}[!ht]
             \begin{center}
             \includegraphics[height=6cm, width=9cm]{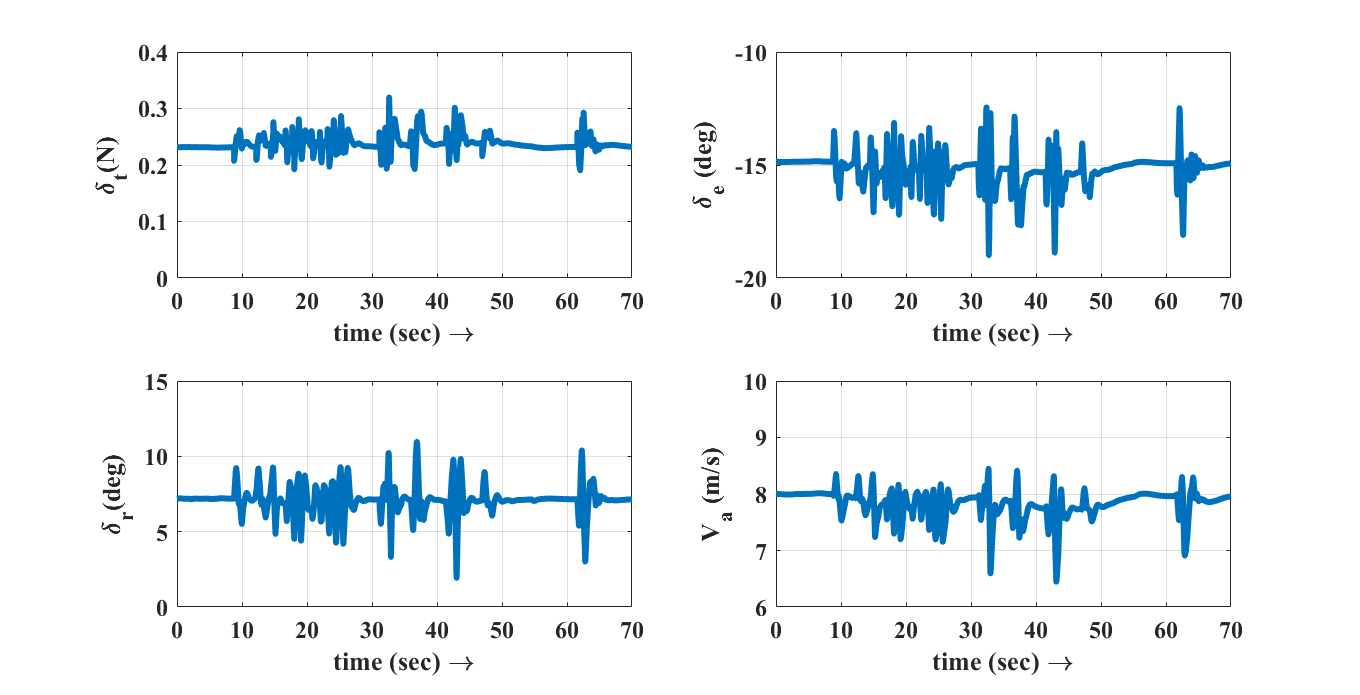}
             \caption{Control inputs and airspeed for rectangular path following}
             \label{inpus2}
             \end{center}
        \end{figure}

%\clearpage
\noindent The objective of the second simulation is to follow a rectangular path. Following a rectangular path enables the MAV to perform surveillance over a given area. The rectangular path considered is of  size  100m x 200m.  The path followed by the MAV is shown in Fig. \ref{xyplots2}. In  Fig. \ref{xyplots2}, small red circles denotes the waypoints along the rectangular path and blue line represents the path followed by the MAV. The starting point of the flight path is at (0,0). The MAV takes a turn after reaching the first waypoint at (80,0). The plot of $d$ and $\rho$ given in Fig. \ref{missds2} show oscillations whenever MAV switches to a new waypoint. The miss distance goes to zero as the MAV approaches a waypoint. The plot of control inputs and airspeed for rectangular path following is shown in Fig. \ref{inpus2}. The airspeed is between 7-8.5 $m/s$ for most of the time duration of the MAV flight.

\section{Conclusion}
\noindent The paper presents a novel waypoint navigation algorithm in IGC framework. The  IGC methodology combines unstable MAV coupled dynamics with the pure proportional navigation guidance law. The resulting linear closed loop system is found to be stable with a static output feedback control law. The proposed waypoint navigation algorithm handles the minimum turn radius constraint of the MAV and determines the feasibility of reaching a given waypoint. The waypoint navigation algorithm does not require any iterative methods to compute the control inputs.  The high fidelity nonlinear numerical simulations for straight line and rectangular path following validates the utility of the algorithm. The airspeed of the MAV is maintained well above the stall value while performing the path following. The extension of this algorithm for waypoint navigation in three-dimensional space would be  interesting future work.

\appendices

\section*{Acknowledgment}

\noindent The authors would like to thank National Programme for Micro Air Vehicle (NPMICAV) and Aeronautical Development and Research Board (ARDB), government of India for funding the project.

\ifCLASSOPTIONcaptionsoff
  \newpage
\fi


\begin{thebibliography}{1}

\bibitem{mav1} K. Harikumar, J. V. Pushpangathan, T. Bera, S. Dhall, and M. S. Bhat. "Modeling and closed loop flight testing of a fixed wing micro air vehicle," Micromachines,  Vol. 9, No. 3, p.111, 2018.

  
  \bibitem{Mclain} D.R. Nelson, D.B. Barber, T.W. McLain  and R.W. Beard, "Vector field path following for miniature air vehicles," IEEE Transactions on Robotics, Vol. 23, No. 3, pp. 519-529, 2007.


\bibitem{hari1} K. Harikumar, Sidhant Dhall and M. Seetharama Bhat, "Nonlinear modeling and control of coupled dynamics of a fixed wing micro air vehicle," Indian Control Conference, Hyderabad, India, January  2016.

\bibitem{shika} Sikha Hota and Debasish Ghose, "Time-Optimal Convergence to a Rectilinear Path
in the Presence of Wind," Journal of Intelligent Robotic Systems  Vol. 74, pp. 791-815, 2014.

\bibitem{hari2}{K. Harikumar}, Sidhant Dhall and M. Seetharama Bhat, "Design and experimental validation of a Robust output feedback control
for the coupled dynamics of a micro air vehicle," \textit{{Springer International Journal of Control, Automation and Systems}}, vol. 17, no. 1, pp. 155-167, 2019.







\bibitem{Hyan}  H. Yan and H. Ji, "Integrated guidance and control for dual-control missiles based on
small-gain theorem," Automatica Vol. 48, pp. 2686-2692, 2012.
\bibitem{Mxin} M. Xin, S. N. Balakrishnan and E. J. Ohlmeyer, "Integrated Guidance and Control of Missiles
With $\theta$- D Method," IEEE Transactions on  Control Systems Technology, Vol. 14, No. 6, pp. 981-992, 2006.
\bibitem{Yike} Y. Ke, T. Qingke, L. Siyuan, L. Qingdong and R. Zhang, "Integrated guidance and control design based on motion tracking," Proceedings of the 35th Chinese Control Conference, Chengdu, China, pp. 5721-5725, 2016.



 \bibitem{menon} P.K. Menon and E.J. Ohlmeyer, "Integrated design of agile missile guidance and autopilot systems," Control Engineering Practice, Vol. 9, No. 10, pp. 1095-1106, 2001.
 
 \bibitem{isa1} C. Zhang and W. Yun-jie, "Non-singular terminal dynamic surface control based integrated guidance and control design and simulation," ISA transactions, Vol. 63, pp. 112-120, 2016.
 
 \bibitem{ast1} G. Jianguo, Y. Xiong, and J. Zhou, "A new sliding mode control design for integrated missile guidance and control system," Aerospace Science and Technology,  Vol.78, pp. 54-61, 2018.

\bibitem{kaminer} I. Kaminer, A. Pascoal, E. Hallberg and C. Silvestre, "Trajectory tracking for autonomous vehicles: an integrated approach to guidance and control," Journal of Guidance, Control and Dynamics, Vol. 21, No. 1, pp. 29-38, 1998.

\bibitem{yamasaki} T. Yamasaki, S.N. Balakrishnan and H. Takano, "Separate channel integrated guidance and control for automatic path following," Journal of Guidance, Control and Dynamics, Vol. 36, No. 1, pp. 25-34, 2013.

\bibitem{padhi} R. Padhi, P. R. Rakesh and R. Venkataraman, "Formation Flying with Nonlinear Partial Integrated Guidance and Control," IEEE Transactions on Aerospace and Electronic Systems, Vol. 50, No. 4, pp. 2847-2859, 2014.
\bibitem{xliu} X. Liu, W. Huang, L. Du , P. Lan and Y. Sun, "Three-Dimensional Integrated Guidance and Control for BTT Aircraft Constrained by Terminal Flight Angles, " 27th Chinese Control and Decision Conference (CCDC), pp.107-112, 2015. 

\bibitem{babar} M. Z. Babar, R. Samar, A. I. Bhatti and M. Baglietto, "Robust Integrated Lateral Guidance and Control of
UAVs," 20th IEEE International Conference on  Methods and Models in Automation and Robotics (MMAR), pp. 523-528, 2015.

\bibitem{jj} J. V. Pushpangathan, M. S. Bhat, and K. Harikumar, "Effects of
Gyroscopic Coupling and Countertorque in a Fixed-Wing Nano Air
Vehicle," Journal of Aircraft, vol. 55, no.1, pp. 239-250,  2018.

 

\bibitem{Roskam}J. Roskam, "Airplane design," Part 6, Roskam Aviation and Engineering Corporation, pp. 371-461, 1990.




\bibitem{shneydor} N.A. Shneydor, "Missile Guidance and Pursuit: Kinematics, Dynamics and control," Horwood publishing, England, pp. 109-117, 1998.

\bibitem{yansh} R. Yanushevsky, "Guidance of Unmanned Aerial Vehicles," CRC Press, pp. 9-18,
2011.
\bibitem{meriam} J.L. Meriam, L.G. Kraige, "Engineering mechanics - Dynamics," Sixth edition, Wi-
ley India, pp. 55-57, 2008.
\bibitem{jiankun} H. Jiankun, C. Bohn and H.R. Lu, "Systematic $H_{\infty}$ weighting function selection
and its application to the real-time control of a vertical take-off aircraft," Control Engineering Practice, Vol. 8, No. 3, pp. 241-252, 2000.


\end{thebibliography}
\end{document}